\documentclass[10pt,twocolumn,letterpaper]{article}

\usepackage[pagenumbers]{iccv}              

\usepackage[dvipsnames]{xcolor}
\usepackage{soul} 
\usepackage{lipsum}
\usepackage{xcolor}

\definecolor{gray}{rgb}{0.7,0.7,0.7}
\definecolor{lightblue}{rgb}{0.3, 0.5, 1}
\definecolor{lightgreen}{rgb}{0.1, 0.5, 0.1}
\definecolor{salmon}{rgb}{1, 0.3, 0.3}

\usepackage{graphicx}
\usepackage{amsmath}
\usepackage{amssymb}  
\usepackage{lipsum}
\usepackage{multirow}

\usepackage{amsthm}
\newtheorem{definition}{Definition}
\newtheorem{thm}{Theorem}
\newtheorem{corollary}{Corollary}[thm]

\usepackage[ruled]{algorithm}
\usepackage{algpseudocode}

\usepackage{rpm_SIunits}
\usepackage{rpm_acronyms}
\usepackage{rpm_math}
\usepackage{rpm_misc}

\usepackage{subcaption}

\usepackage{amsthm}
\usepackage[accsupp]{axessibility}
\usepackage{adjustbox}
\newtheorem{problem}{Problem}
\definecolor{iccvblue}{rgb}{0.21,0.49,0.74}
\newcommand{\iccvbl}[1]{{\textcolor{iccvblue}{#1}}}

\definecolor{iccvblue}{rgb}{0.21,0.49,0.74}
\usepackage[pagebackref,breaklinks,colorlinks,allcolors=iccvblue]{hyperref}

\def\paperID{9340} 
\def\confName{ICCV}
\def\confYear{2025}

\title{

Registration beyond Points: General Affine Subspace \\ Alignment via Geodesic Distance on Grassmann Manifold

}

\author{Jaeho Shin${}^{1}$ \;\; Hyeonjae Gil${}^{1}$ \;\; Junwoo Jang${}^{2}$ \;\; Maani Ghaffari${}^{3}$ \;\; Ayoung Kim${}^{1\dag}$\\
${}^{1}$Seoul National University \;\; ${}^{2}$Inha University \;\; ${}^{3}$University of Michigan\\}

\begin{document}
\maketitle
\begin{abstract}

Affine Grassmannian has been favored for expressing proximity between lines and planes due to its theoretical exactness in measuring distances among features. Despite this advantage, the existing method can only measure the proximity without yielding the distance as an explicit function of rigid body transformation. Thus, an optimizable distance function on the manifold has remained underdeveloped, stifling its application in registration problems. This paper is the first to explicitly derive an optimizable cost function between two Grassmannian features with respect to rigid body transformation ($\mathbf{R}$ and $\mathbf{t}$). Specifically, we present a rigorous mathematical proof demonstrating that the bases of high-dimensional linear subspaces can serve as an explicit representation of the cost. Finally, we propose an optimizable cost function based on the transformed bases that can be applied to the registration problem of any affine subspace. Compared to vector parameter-based approaches, our method is able to find a globally optimal solution by directly minimizing the geodesic distance which is agnostic to representation ambiguity. The resulting cost function and its extension to the inlier-set maximizing \ac{BnB} solver have been demonstrated to improve the convergence of existing solutions or outperform them in various computer vision tasks. The code is available on \url{https://github.com/joomeok/GrassmannRegistration}.
 

\end{abstract}

\section{Introduction}
\label{sec:intro}

Registration is a fundamental problem in computer vision, SLAM, and object pose estimation \cite{CVPR-2016-Schonberger, TRO-2015-Mur}. While deep feature-based methods have become increasingly popular, geometric registration remains essential for its interpretability and efficiency. Among these, 3D point-based methods \cite{RSS-2009-Segal, Review-2015-Pomerleau} are widely adopted but fail to capture structural relationships and are highly sensitive to sensor noise.


\begin{figure}[!t]
    \centering
    \includegraphics[width=1\linewidth]{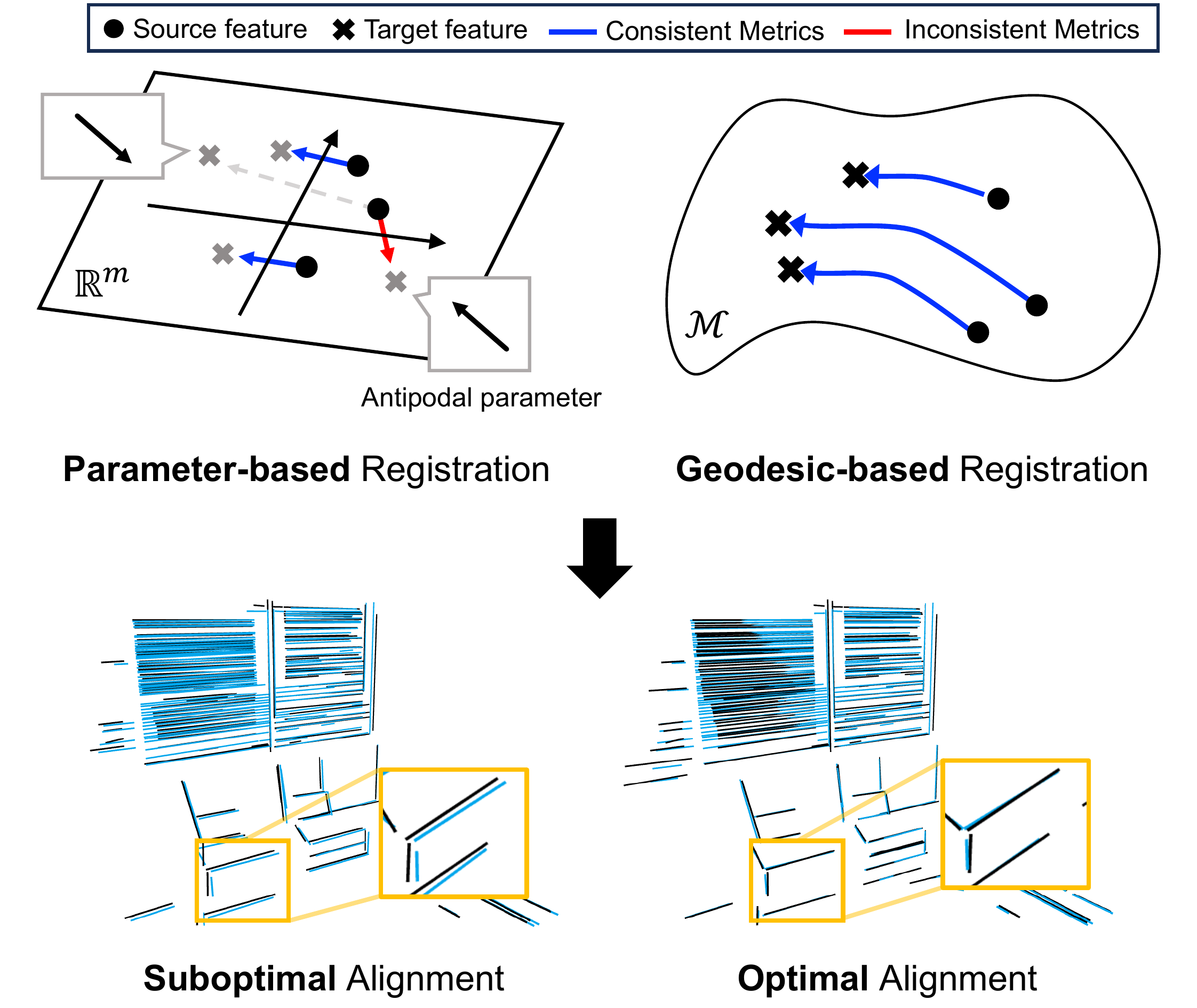} 
    \caption{\textbf{Comparison of parameter and geodesic-based registration.} Existing methods for solving line and plane registration minimize the squared sum of distances between vectors used to parameterize features in Euclidean space (\eg, Plücker coordinates, plane parameters). However, the sign ambiguity of vector representations induces metric inconsistencies and suboptimality. In contrast, as geodesic distances represented by basis span is agnostic to such ambiguity, our cost function provides the optimal registration result.
    }
    \label{fig:Fig1}
    \vspace{-5mm}
\end{figure}


To overcome these limitations, higher-order geometric features such as lines and planes \cite{TRO-2019-Gomez, CVPR-2018-Huang, CVPR-2023-Liu} have emerged as a robust alternative, as they naturally encode scene structure with noise resilience. However, many approaches either approximate feature distances using point-based approximation \cite{ICPR-2008-Olsson, PAMI-2008-Olsson, CVPR-2017-Briales} or parameterize them naively in Euclidean space \cite{ICCVW-2017-forstner, CVPR-2021-Liu}. These approaches can reintroduce noise sensitivity or lead to inconsistent metrics due to potential sign ambiguities (\eg, a plane’s normal vector and its negative representing the same plane), ultimately resulting in suboptimal registration accuracy, as illustrated in \figref{fig:Fig1}.

Alternatively, lines and planes can be represented as affine subspaces (defined by a linear subspace and an offset) with their formulation on the affine Grassmannian \cite{IROS-2022-Lusk, arxiv-2024-Yu}. This interpretation enables a principled definition of the Grassmann distance, a natural metric for geometric proximity that quantifies differences in orientation and position between subspaces. While this approach eliminates issues in point-based and naive Euclidean representations, leveraging these advantages for registration requires an optimizable cost function parameterized by rotation $\mathbf{R}$ and translation $\mathbf{t}$, which has remained unexplored so far.

To address this problem, we derive novel properties of the affine Grassmannian in the context of registration and propose a general framework for solving affine subspace alignment. First, we derive an explicit formulation of the orthogonal displacement vector under rigid transformations. Then, we show that the bases of a linear subspace provide an exact intermediate representation for distance measurement on the manifold, enabling the construction of an optimizable cost function. To the best of our knowledge, this is the first work to use the affine Grassmannian and its distance metric for subspace registration. Our contributions are summarized as follows:

\begin{enumerate}
    \item We propose an optimizable cost function for general line and plane registration that minimizes the geodesic distance on the Grassmann manifold. With consistent metrics across correspondences, our cost significantly improves the convergence of existing linear solvers.

    \item In cases where many outliers or unknown correspondences are present, we implement an inlier-set maximization \ac{BnB} solver by deriving exact bounds from our cost function for 3D \textit{line-to-line}, \textit{line-to-plane}, and \textit{plane-to-plane} registration.
    \item The proposed algorithm is applied to various computer vision problems, including object registration, RGB-D odometry, \ac{PnL}, and localization, and  demonstrates superior performance on each task compared to previous approaches.
\end{enumerate}


\section{Related Works}
\label{sec:related_work}

This section briefly reviews two lines of studies in the existing line and plane registration, followed by a summary of relevant literature on the Grassmann manifold as a useful structure for embedding subspace data.



\noindent\textbf{Line and Plane Registration in $\mathbb{R}^3$.} Many algorithms on registration with lines and planes incorporate points as an intermediate feature and solve for \textit{point-to-line} and \textit{point-to-plane} distances, which should be zero when the given point lies on the feature. One approach for obtaining the solution from this constraint is marginalizing translation and solving the constrained problem by $SO(3)$ condition exploiting Lagrangian relaxation of the primal problem \cite{ICPR-2008-Olsson, PAMI-2008-Olsson, CVPR-2017-Briales}. Similarly, \citet{RAL-2023-Malis, RAL-2024-Malis} derived a closed-form solution from polynomials of quaternion by the Lagrange multiplier method. 

Another approach exploits line or plane parameters and minimizes Euclidean distance between the associated parameters (\eg, Plücker coordinate or plane coefficients). Adopting rotation estimation strategy as in \cite{PAMI-1991-Chen}, \citet{CVPR-2021-Liu} solved a 2-line minimal solver with RANSAC scheme by directly solving equality of Plücker coordinate. For registering planes, \citet{ICCVW-2017-forstner} solved maximum-likelihood estimation from the transformation rule of the plane parameter, considering its uncertainty from the point cloud. 
By limiting the high-dimensional feature in an approximated vector space, this heuristic parameterization and the distance metric depend on the design choice, failing to represent the exact adjacency among features.



\noindent \textbf{Grassmann Manifold}. Grassmann manifold has been extensively used in various research areas of computer vision, mainly serving as an alternative space for analyzing data that can be treated as a linear subspace. Tasks such as face recognition \cite{CVPR-2015-Huang}, object tracking \cite{WACV-2014-Shirazi}, and \ac{SFM} \cite{CVPR-2018-Kumar} have been addressed with metrics on the manifold to determine the similarity between images. Recent studies on geometric deep learning also utilized the advantage of this space for handling the subspace data. \citet{ICLRW-2021-Zhou} embedded multiple graphs represented as graph convolutional network into an element on the manifold and performed graph classification tasks. \citet{NIPS-2023-Yataka} also showed that a continuous flow generation model can be applied for learning distribution on Grassmann manifold and generated various 1-dimensional shapes. An investigation by \citet{IROS-2022-Lusk} most closely aligns with our approach to handle 3D lines and planes as affine subspace. 
However, their approach is limited in using the geodesic distance only when constructing the consistency graph and relied on parameters to restore relative pose. 

\section{Preliminary}
\label{sec:preliminary}

\subsection{Notations}\label{Pre:Notation}

We denote scalars using italicized characters, vectors with boldface lowercase characters, matrices with boldface uppercase characters, and subspaces with blackboard bold characters. Given a $n$-dimensional vector $\mathbf{v}$, we define $\bar{\mathbf{v}}$ as $(n+1)$-dimensional vector by augmenting the $\mathbf{v}$ with $0$ at the last element. Similarly, $\tilde{\mathbf{v}}$ is $(n+1)$-dimensional vector by augmenting $\mathbf{v}$ with $1$ and normalizing it to norm-1, which is $\tilde{\mathbf{v}} = [\frac{\mathbf{v}^\top}{\sqrt{1+\|\mathbf{v}\|^2}},\frac{1}{\sqrt{1+\|\mathbf{v}\|^2}}]^\top$. Also, $\left\|\mathbf{v}\right\|_2$ denotes Euclidean norm of the vector $\mathbf{v}$. Given a linear subspace $\mathbb{U} \subseteq  \mathbb{R}^n$ with its orthonormal basis matrix $\mathbf{U}$, $\mathbf{P}_\mathbb{U} := \mathbf{U}\mathbf{U}^\top \in \mathbb{R}^{n \times n}$ denotes the projection matrix which maps any vector $\mathbf{x}\in\mathbb{R}^n$ onto its projection $\mathbf{P}_\mathbb{U}\mathbf{x}$ within $\mathbb{U}$. 

\subsection{Affine Grassmannian}\label{Pre:Graff}

We review the concept of Grassmannian and its corresponding manifold on affine subspaces before stating our algorithm. The Grassmannian $\mathrm{Gr}(k,n)$ is defined as the set of all $k$-dimensional linear subspaces of the $n$-dimensional Euclidean space. For example, a 3D line passing through the origin can be interpreted as the element of $\mathrm{Gr}(1,3)$, since it is a 1-dimensional linear subspace of $\mathbb{R}^3$. An element of Grassmannian $\mathbb{V} \in \mathrm{Gr}(k,n)$ can also be represented by its $k$ orthonormal basis, stacked column-wise and forming the orthonormal matrix representation by $\mathbf{Y}_\mathbb{V}$. Inducing the geodesic distance on this manifold, known as Grassmann distance, is elaborated in \apref{Appendix:GrassDist}. For two 1-dimensional subspaces, we express Grassmann distance as follows, which is an acute angle formed by two bases:

\begin{definition}[1D Grassmann Distance]
Grassmann distance between 1D subspaces spanned by $\mathbf{u}$ and $\mathbf{v}$ is defined as:
\begin{equation}\label{equ:1D_dist}
\mathrm{d}_{\mathrm{Gr}}\left(\mathrm{span}\{\mathbf{u}\},\mathrm{span}\{\mathbf{v}\}\right) = \cos^{-1}(\left|\mathbf{u}^\top \mathbf{v} \right|).
\end{equation}
\end{definition}
\noindent Throughout this paper, we denote linear subspace spanned by a 1-dimensional vector omitting the symbol $\mathrm{span}\{\}$ for convenience. Extending the concept of the manifold to broader subspaces, we derive a natural generalization of the Grassmannian of affine subspaces:

\begin{definition}[Affine Grassmannian]
\label{def:GrAff}
Let  $k < n$ be positive integers. The set of all k-dimensional affine subspaces of $\mathbb{R}^n$ is defined as affine Grassmannian, denoted by $\mathrm{Graff}(k,n)$.
\end{definition}
A $k$-dimensional affine subspace, denoted as \mbox{$\mathbb{A}+\mathbf{b} \in \mathrm{Graff}(k, n)$}, consists of a linear subspace $\mathbb{A} \in \mathrm{Gr}(k, n)$ and a displacement vector $\mathbf{b} \in \mathbb{R}^n$ from the origin. 
This affine subspace can be represented by an orthonormal matrix $\mathbf{A}$ and a vector $\mathbf{b}_0$, where the column vectors of $\mathbf{A}$ form the orthonormal basis of the linear subspace, and $\mathbf{b}_0$ is a unique displacement vector orthogonal to $\mathbf{A}$. The detailed process of obtaining the unique displacement vector from an arbitrary displacement is explained in \apref{Appendix:Disp}. The orthonormal matrix representation for the affine Grassmannian is defined from its embedding $z(\mathbb{A}+{\mathbf{b}_0})$, which is an element of Grassmannian in higher dimension \cite{SIAM-2021-Lim}: \begin{definition}[Orthonormal Matrix Representation of Affine Grassmannian]
\label{def:GraffCoord}
The orthonormal basis matrix for $\mathbb{A}+\mathbf{b}_0 \in\mathrm{Graff}(k,n)$ is the $(n+1)\times(k+1)$ matrix defined as:
\begin{equation}
\mathbf {Y}_{z({\mathbb{A}+\mathbf{b}_0})} =    
\begin{pmatrix}
\mathbf{A} & \frac{\mathbf{b}_0}{\sqrt{1+\|\mathbf{b}_0\|^2}} \\
0 & \frac{1}{\sqrt{1+\|\mathbf{b}_0\|^2}} \\
\end{pmatrix}.
\end{equation}

\end{definition}
\noindent \defref{def:GraffCoord} enables us to obtain the geodesic distance between affine subspaces, as the affine Grassmannian inherits the Grassmann distance in the higher dimensional space as its metric.

\section{Methodology}
\label{sec:method}


\subsection{Motion in Affine Primitives}

\begin{figure*}[!t]
    \centering
    \includegraphics[width=0.8\linewidth]{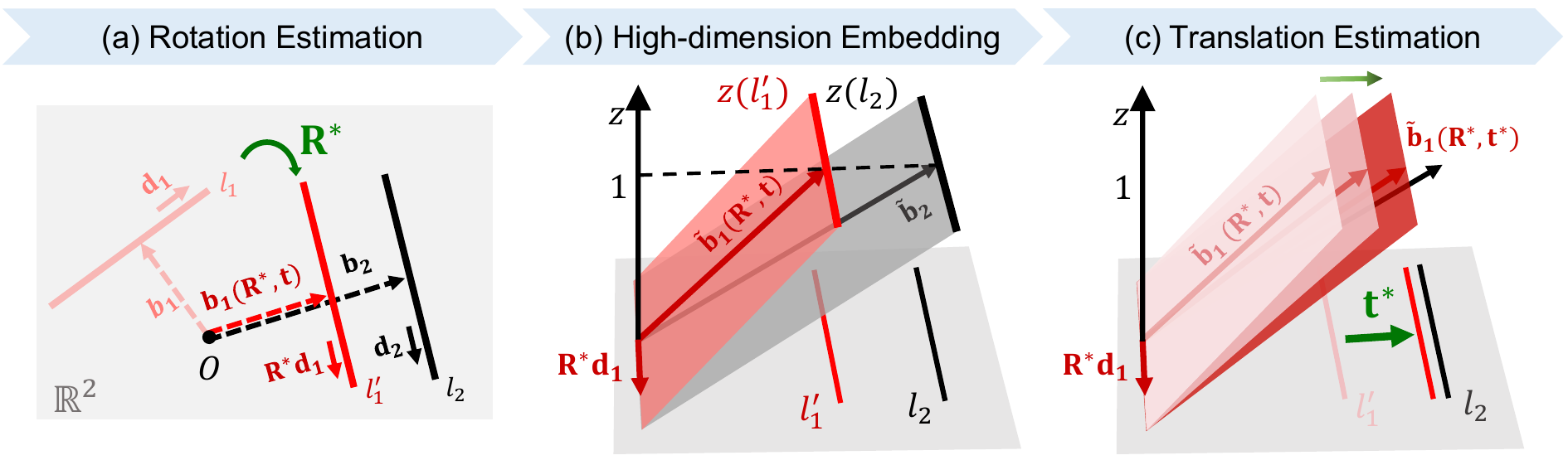}
    \caption{\textbf{Registration of two 2D lines by separate estimation}. (Left) Given a pair of 2D lines ($l_1$, $l_2$), our inlier-set maximization algorithm estimates the rotation matrix by making each basis of linear subspace aligned, which is a 2D direction vector. (Middle, Right) After obtaining the rotation, a translation vector is estimated by inducing the secondary basis vector of the embedded subspace $z(l_2)$, denoted as $\tilde{\mathbf{b}}_2$, to be included in the transformed and embedded line $z(l_1')$. This is achieved by minimizing the distance between $\tilde{\mathbf{b}}_2$ and its projection onto $z(l_1')$ to zero. For clarity, $\tilde{\mathbf{b}}_1$ and $\tilde{\mathbf{b}}_2$ are illustrated to have length longer than 1.} \vspace{-4mm}\label{fig:2DExample}
\end{figure*}

To handle the registration of affine primitive under rotation and translation in $\mathbb{R}^n$, we first define the group of rigid transformation as follows:
\begin{equation}
     \textit{SE(n)} = \left\{\begin{bmatrix}
\mathbf{R} & \mathbf{t} \\
\mathbf{0} & 1 \\ \end{bmatrix} \in GL_{n+1}(\mathbb{R}) \mid \mathbf{R} \in SO(n), \mathbf{t}\in\mathbb{R}^n \right\}, \nonumber
\end{equation} where $GL_{n+1}(\mathbb{R})$ is the general linear group of degree ($n+1$) over $\mathbb{R}$ and $SO(n)$ is the $n$-dimensional special orthogonal group. 
For any $\mathbf{R} \in SO(n)$ and $\mathbb{A} \in \mathrm{Gr}(k,n)$ with its orthonormal matrix representation $\mathbf{A}$, the action by $\mathbf{R}$ on $\mathrm{Gr}(k,n)$ yields:

\begin{equation}
    \mathbf{R} \cdot \mathbb{A} := \mathrm{span}\{\mathbf{RA}\} \in \mathrm{Gr}(k,n). \nonumber
\end{equation} Extending this, we obtain the following group action of $SE(n)$ on affine Grassmannian: 

\begin{thm}[\textit{SE(n)} Action on Affine Grassmannian]\label{thm:SE(n)Action}
A function {\small$f: SE(n) \times \mathrm{Graff}(k,n) \rightarrow \mathrm{Graff}(k,n)$} defined as:
{
\begin{align}\label{equ:SE(n)Action}
    \mathbf{T} \cdot (\mathbb{A}+\mathbf{b}_0) &= (\mathbf{R} \cdot \mathbb{A}) + (\mathbf{Rb}_0 + \mathbf{R}(\mathbf{I} - \mathbf{A}\mathbf{A}^\top)\mathbf{R}^\top\mathbf{t}) \\
    &=  (\mathbf{R} \cdot \mathbb{A}) + \mathbf{b}_0'(\mathbf{R},\mathbf{t}),
\end{align}}
is a group action, where orthonormal basis matrix leads to:
\begin{equation}
     \mathbf{Y}_{z(\mathbf{T} \cdot (\mathbb{A}+\mathbf{b}_0))} = \begin{bmatrix}
\mathbf{RA} & \frac{\mathbf{b}_0'(\mathbf{R},\mathbf{t})}{\sqrt{1+\left\|\mathbf{b}_0'(\mathbf{R},\mathbf{t}) \right\|^2}} \\
\mathbf{0} & \frac{1}{\sqrt{1+\left\| \mathbf{b}_0'(\mathbf{R},\mathbf{t})\right\|^2}} \\
\end{bmatrix}.
\end{equation}
\end{thm}
\begin{proof}
    See \bl{\apref{Appendix:Theorem1}}. 
\end{proof}
\noindent We now prove the compatibility of this action with standard group action of $SE(n)$ on Euclidean space as follows:

\begin{corollary}[] \label{cor:EuclideanAction}
For all point $\mathbf{x}$ included in an affine subspace $\mathbb{A}+\mathbf{b}_0$, $\mathbf{T}\cdot\mathbf{x}$ is also included in $\mathbf{T}\cdot(\mathbb{A} + \mathbf{b}_0)$, where $\mathbf{T} \cdot \mathbf{x}:=\mathbf{R}\mathbf{x}+\mathbf{t}$ is a transformed point by the standard $SE(n)$ action on $\mathbb{R}^n$. 
\end{corollary}
\begin{proof}
    See \bl{\apref{Appendix:Corollary1.1}}. 
\end{proof}
\noindent This allows us to reformulate registration problems in Euclidean space using the affine Grassmannian, where group action on the manifold represents the transformation of subspace by $\mathbf{T}\in SE(n)$ in $\mathbb{R}^n$, implying if two affine subspaces on Euclidean space are registered by $\mathbf{T}$, they also have zero distance on the manifold. 

\subsection{General Affine Primitive Registration}\label{sec:AffineRegist} So far, we have derived how the components of elements on the manifold change under rigid transformation in Euclidean space. Exploiting this result, we now explain how to obtain an optimizable distance function between the transformed source and target subspaces for registration. First, let us define the primal problem that aims to estimate $\mathbf{T}\in SE(n)$ minimizing the squared sum of geodesic distance on the manifold. For clarity, we assume correspondences are given as prior information.

\begin{problem}\label{problem:original} Let $k$, $l$ and $n$ be positive integers satisfying $k \leq l < n$. The registration problem between $N$ paired affine primitives $(\mathbb{A}^i+\mathbf{c}_0^i,\mathbb{B}^i+\mathbf{d}_0^i)\in \mathrm{Graff}(k,n)\times\mathrm{Graff}(l,n)$, ($i=1,\cdots, N$) is defined by estimating the minimizer of the squared sum of Grassmann distance between embedded subspaces, which is: 
\begin{equation}\label{equ:Problem1}
 \mathrm{arg}\min_\mathbf{T}\sum^N_{i=1} \mathrm{d}_\mathrm{Gr}\left(z(\mathbb{A}^i+\mathbf{c}_0^i),z(\mathbf{T} \cdot (\mathbb{B}^i+\mathbf{d}_0^i))\right)^2.   
\end{equation}
\end{problem}

\noindent This formulation can be widely adopted in aligning arbitrary geometric features and works even when two subspaces have different dimensions (\eg, between line and plane). 
However, as can be seen, \equref{equ:Problem1} is not explicitly expressed in terms of $\mathbf{R}$ and $\mathbf{t}$, making it unsuitable for direct application to the registration problem. Additionally, the Grassmann distance is obtained from \ac{SVD} \cite{bjorck1973numerical}, which is not differentiable with respect to $SE(n)$. To bridge this gap, we derive the following theorem, which allows us to equivalently optimize \equref{equ:Problem1} by introducing the basis of the embedded subspace as an intermediate representation.


\begin{thm}[Equivalence of Zero Grassmann Distance]
\label{thm:GrasEqui}
The Grassmann distance between two linear subspaces is zero if and only if every orthonormal basis of a smaller-dimensional subspace is spanned by the orthonormal bases of another subspace.
\end{thm}
\begin{proof}
    See \bl{\apref{Appendix:Theorem2}}.
\end{proof}

\thmref{thm:GrasEqui} proves that the discrepancy between the bases and their projection onto another subspace can function as an indicator to determine the proximity of two features. From this theorem, Problem \ref{problem:original} can now be alternatively optimized since bases are directly expressed with $\mathbf{R}$ and $\mathbf{t}$. Also, as the comparison with its projection is agnostic to the direction of the basis, ambiguity related to feature representation can now be disregarded. As a result, we obtain the optimizable reformulation of Problem \ref{problem:original} as follows:



\begin{problem}[Affine Primitives Registration]\label{problem:redefine}
The registration between affine primitives in Problem \ref{problem:original} is explicitly redefined as:
{\scriptsize
\begin{equation}\label{equ:Problem2}
\mathrm{arg}\min_{\mathbf{T}}\sum_{i=1}^{N}\left(\sum_{j=1}^{k}\left\|\mathbf{P}_{\mathbf{R} \cdot \mathbb{B}^i}\mathbf{a}^i_j-\mathbf{a}^i_j\right\|_2^2 +
\left\|\mathbf{P}_{z(\mathbf{T} \cdot (\mathbb{B}^i+\mathbf{d}_0^i))}\tilde{\mathbf{c}}_0^i- \tilde{\mathbf{c}}_0^i\right\|_2^2\right),
\end{equation}
} where $\mathbf{a}^i_j$ denotes $j$th column vector of $\mathbf{A}^i$. 


\end{problem}
\begin{proof}
    See \bl{\apref{Appendix:Problem2}} 
\end{proof}

Indicated by the $(\tilde{\cdot})$ operation, \equref{equ:Problem2} seeks to find the transformation that minimizes the distance between the basis of the embedded subspace and its projection. Detailed process for obtaining the optimizable cost function from raw measurement is given as \cref{alg:OptCost}. In addition to its exactness, this formulation allows us to handle existing registration problems in a unified perspective (\eg, a point is 0-dimensional affine primitive). Also, summation terms for index $j$ in \equref{equ:Problem2} represent the alignment between linear subspace, which is solely dependent on the rotation \textit{SO(n)}. This implies the rotation can be obtained separately from the translation vector. Our theorems also provide a mathematical foundation for previous works estimating rotation from directional information of subspaces \cite{PAMI-1991-Chen, ICCV-2015-Brown, RAL-2020-Liu}. 

\begin{algorithm}[t]
    \footnotesize

    \caption{Optimizable Geodesic Distance Minimization}
    \label{alg:OptCost}
    \hspace*{\algorithmicindent} \textbf{Input:} \\
   \hspace*{\algorithmicindent}\hspace{1em} $\mathcal{X} = \{(\textbf{A}^i,\textbf{c}_0^i)\},\; (i=1,\cdots, N)$: $n_t$-dim target features in $\mathbb{R}^n$\\
   \hspace*{\algorithmicindent}\hspace{1em} $\mathcal{Y} = \{(\textbf{B}^i,\textbf{d}_0^i)\},\; (i=1,\cdots, N)$: $n_s$-dim source features in $\mathbb{R}^n$
    
\hspace*{\algorithmicindent}	\textbf{Output:}\\ \hspace*{\algorithmicindent}\hspace{1em} $f(\mathbf{R},\mathbf{t})$: Total objective cost \quad\text{\% \equref{equ:Problem2}}
    \begin{algorithmic}[1]
    \State $f(\mathbf{R},\mathbf{t})=0$, $\mathcal{Y}_{lin}$ = \{\} 

    \For{$i = 1 : N$} 
        \For{$j=1:n_s$}
            \State {\scriptsize $\mathcal{Y}_{lin}[i][1:n+1,j] \leftarrow \left[(\mathbf{R}{\mathbf{b}}^i_j)^\top,0\right]^\top$}
        \EndFor
            \State {\scriptsize $\mathcal{Y}_{lin}[i][1:n+1,n_s+1] \leftarrow \texttt{Tilde}(\mathbf{R} {\mathbf{d}}^i_{0} + \mathbf{R}(\mathbf{I}-\mathbf{B^i B^i}^\top)\mathbf{R}^\top\mathbf{t})$}

            \For{$j=1:n_t$}
            \State $\mathbf{P}_{\mathbf{R}\cdot\mathbb{B}^i}=\mathcal{Y}_{lin}[i][1:n,1:n_s]\,\mathcal{Y}_{lin}[i][1:n,1:n_s]^\top$
            \State {\footnotesize$f(\mathbf{R},\mathbf{t})  \mathrel{+}= 
            \left\|\mathbf{P}_{\mathbf{R} \cdot \mathbb{B}^i}\mathbf{A}^i[1:n,j]-\mathbf{A}^i[1:n,j]\right\|_2^2$}

            \EndFor
            \State $\mathbf{P}_{z(\mathbf{T} \cdot (\mathbb{B}^i+\mathbf{d}_0^i))}=\mathcal{Y}_{lin}[i] \,\mathcal{Y}_{lin}[i]^\top$
            \State {\footnotesize $f(\mathbf{R},\mathbf{t})  \mathrel{+}=\left\|\mathbf{P}_{z(\mathbf{T} \cdot (\mathbb{B}^i+\mathbf{d}_0^i))}\tilde{\mathbf{c}}_0^i- \tilde{\mathbf{c}}_0^i\right\|_2^2$}           
    \EndFor
    \State \textbf{return} $f(\mathbf{R},\mathbf{t})$
    \end{algorithmic}
\end{algorithm}

\subsection{Solutions for 3D Registration Problem}\label{Methodology:BnB}
As a result of \cref{sec:AffineRegist}, we can now optimize distances between Grassmannian features in terms of $\mathbf{R}$ and $\mathbf{t}$ by using the basis spanning condition (\thmref{thm:GrasEqui}). In this section, we propose specific applications of our cost function for various registration problems in the field of computer vision. First, in the case of \textit{line-to-line} and \textit{plane-to-plane} registration, where typical parameter-based linear solvers exist \cite{CVPR-2021-Liu, ICCVW-2017-forstner}, \equref{equ:Problem2} can be used as the total cost for the subsequent refinement of the solution. However, in this case, the accuracy of the solution may be heavily dependent on the performance of the linear solver in filtering out outlier correspondences.

Therefore, to handle practical cases of noisy correspondences, we propose a formulation for inlier-set maximization \cite{PAMI-2008-Olsson, PAMI-2018-Campbell}, which can be easily derived from \equref{equ:Problem2}. As mentioned in \cref{sec:AffineRegist}, the cost function can be divided into two parts: terms solely related to $\mathbf{R}$ and terms related to both $\mathbf{R}$ and $\mathbf{t}$. For clarity, we denote each term for the $i$th correspondence as $f_i$ and $g_i$, respectively. For example, in \textit{line-to-line} registration, $f_1$ represents the term dependent on $\mathbf{R}$ and the direction vectors of the first paired 3D lines. By solving the inlier-set maximization problem only for the $f_i$ terms, we can obtain the optimal rotation $\mathbf{R}$ and the inlier-set correspondences. Specifically, we first solve the following problem:

\begin{equation}\label{equ:rot_min}
  \mathrm{arg}\max_{\mathbf{R}}\sum_{i=1}^{N}\mathbf{1}(\epsilon - f_i(\mathbf{R})),
\end{equation} where $\epsilon$ is inlier threshold and $\mathbf{1}$ refers to the indicator function which outputs the integer $1$ only when the input is a positive number. Then, we subsequently optimize the translation by minimizing the sum of the remaining term $g_i$, using rotation $\mathbf{R}^*$ obtained from the previous step:
\begin{equation}\label{equ:trans_min}
    \mathrm{arg}\min_{\mathbf{t}}\sum_{i=1}^{N}g_i(\mathbf{R}^*,\mathbf{t}).
\end{equation} \noindent Overall pipeline of this two-staged estimation is illustrated in \figref{fig:2DExample} for the case of \textit{line-to-line} alignment.

We also elaborate on $f_i$ and $g_i$ for the three cases of \textit{line-to-line}, \textit{line-to-plane}, and \textit{plane-to-plane} registration. Specifically, to facilitate the optimization process, we utilize the following property of vector projection: the Euclidean distance between a vector and its projection onto a subspace is proportional to the acute angle between the two vectors. As a result, we can replace the Euclidean distance between two vectors in $f_i$ with the one-dimensional Grassmann distance:

\begin{corollary}[Line-to-Line Registration] \label{cor:L2L_dist}
Given $N$ paired 3D lines ($l^i_1$, $l^i_2$), where 
$\mathnormal{l}^i_1 := \mathbf{d}^i_1 + \mathbf{b}^i_1 \in \mathrm{Graff}(1,3)$ and $\mathnormal{l}^i_2 := \mathbf{d}^i_2 +  \mathbf{b}^i_2 \in \mathrm{Graff}(1,3)$, line-to-line registration is defined from the following $f_i$ and $g_i$:
\begin{align}
    f_i(\mathbf{R}) =& \mathrm{d}_\mathrm{Gr}(\mathbf{Rd}^i_1,\mathbf{d}^i_2)^2, \\ 
    g_i(\mathbf{R},\mathbf{t}) =&  \left\|\mathbf{P}_{z(\mathbf{T} \cdot l^i_1)}\tilde{\mathbf{b}}^i_2-\tilde{\mathbf{b}}^i_2\right\|_2^2.
\end{align}

\end{corollary}
\begin{proof}
    See \bl{\apref{Appendix:Cor2}}. 
\end{proof}

\begin{corollary}[Line-to-Plane Registration]\label{cor:L2P_dist}
Given N paired 3D lines and planes ($l^i$,$\pi^i$), where $\mathnormal{l}^i:= \mathbf{d}^i +\mathbf{b}^i\in \mathrm{Graff}(1,3)$ and
$\pi^i:= \mathbb{B}^i + \mathbf{c}^i \in \mathrm{Graff}(2,3)$, line-to-plane registration is defined from the following $f_i$ and $g_i$: 
\begin{align}
    f_i(\mathbf{R}) &= \mathrm{d}_\mathrm{Gr}(\mathbf{d}^i, \mathbf{P}_{\mathbf{R} \cdot \mathbb{B}^i}\mathbf{d}^i)^2,\\ 
    g_i(\mathbf{R},\mathbf{t}) & = \left\|\tilde{\mathbf{b}}^i-\mathbf{P}_{z(\mathbf{T} \cdot \pi^i)}\tilde{\mathbf{b}}^i\right\|_2^2.
\end{align}

\end{corollary}
\begin{proof}
    See \bl{\apref{Appendix:Cor2}}. 
\end{proof}
For the case of \textit{plane-to-plane} alignment, the cost consists of three terms because both subspaces are 2-dimensional affine primitives. We modify the result to yield a compact form of two terms by utilizing the duality of Grassmann manifold: an element $\mathbb{W} \in \mathrm{Gr}(k,n)$ is isomorphic to $\mathbb{W}^\perp \in \mathrm{Gr}(n-k,n)$, which is $(n-k)$-dimensional orthogonal complements of $\mathbb{W}$. Applying the property to $\mathrm{Gr}(2,3)$, we can make two basis vectors of a plane spanned by the other plane bases using the normal vector, which is an orthogonal complement of the 3D plane:

\begin{corollary}[Plane-to-Plane Registration]\label{cor:P2P_dist}
Given $N$ paired 3D planes ($\pi^i_1$, $\pi^i_2$), where $\pi^i_1:= \mathbb{B}^i_1+\mathbf{c}^i_1\in \mathrm{Graff}(2,3)$ with plane normal $\mathbf{n}^i_1$ and $\pi^i_2:= \mathbb{B}^i_2 + \mathbf{c}^i_2 \in \mathrm{Graff}(2,3)$ with plane normal $\mathbf{n}^i_2$, plane-to-plane registration is defined from the following $f_i$ and $g_i$:
\begin{align}
    f_i(\mathbf{R}) &= \mathrm{d}_\mathrm{Gr}(\mathbf{Rn}^i_1, \mathbf{n}^i_2)^2, \\ g_i(\mathbf{R},\mathbf{t}) &= \left\|\mathbf{P}_{z(\mathbf{T} \cdot \pi^i_1)}\tilde{\mathbf{c}}^i_2-\tilde{\mathbf{c}}^i_2\right\|_2^2.
\end{align}
\end{corollary}
\begin{proof}
    See \bl{\apref{Appendix:Cor2}}. 
\end{proof} For solving \equref{equ:rot_min} and \equref{equ:trans_min}, we propose integrating \ac{BnB} as a back-end solver, following numerous prior works that use the algorithm for outlier-robust global optimization \cite{PAMI-2008-Olsson, PAMI-2018-Campbell}. Due to space limitations, upper and lower bounds for each case are provided in the \apref{Appendix:RotBnB} and \ref{Appendix:TransBnB}. The entire pipeline of estimating transformation $\mathbf{R}^*$ and $\mathbf{t}^*$ with \ac{BnB} is summarized in \apref{sec:algorithms} for \textit{line-to-line} case. Our pipeline mostly follows \cite{PAMI-2015-Yang, PAMI-2018-Campbell}; however, we perform a least-squares optimization using the Levenberg-Marquardt (LM) algorithm and automatic differentiation via the Ceres solver \cite{Ceres_Solver} whenever the bound improves upon the optimal cost encountered thus far.

\section{Experiments}\label{sec:Experiments}

 
In this section, we validate our approach by comparing it against other algorithms across various computer vision tasks. First, we apply our \textit{plane-to-plane} registration for the object registration task (\cref{Exp:Space}). Next, we show that our line registration method can improve the trajectory estimation, which requires accurate estimation across multiple frames through the RGB-D odometry experiment (\cref{Exp:RGBDodom}). Also, we solve a fundamental problem of camera pose estimation from line correspondence with our \textit{line-to-plane} registration (\cref{Exp:PnL}). Lastly, we provide another application of our solver, which localizes an RGB-D image within a 3D line map without any given correspondences (\cref{Exp:Localization}).

\subsection{Object Registration}\label{Exp:Space}

In this section, we evaluate the proposed method for an object registration task by using \textit{plane-to-plane} registration. 

\noindent \textbf{Datasets:} We simulated the alignment of a 3D CAD model and noisy point measurements obtained from the sensor. Using the Space Station model provided by \cite{PAMI-2008-Olsson}, 100 points were extracted from each of the 13 planes for \textit{point-to-plane} correspondence with Gaussian noise added to each point ($\sigma^2=0.05^2$). Outlier points were randomly generated within a cuboid with a side length of 20 in the model space that encircles the transformed model. Increasing the outlier ratio of the total correspondences to 50\%, we generated 500 random sets for each ratio. By leveraging the model configuration primarily composed of plane features, we fitted planes from the raw measurements. 

\noindent \textbf{Baselines:} We evaluated our \textit{plane-to-plane} registration approaches in comparison to two conventional approaches, including the convex optimization of point-based method (denoted as Olsson) \cite{ICPR-2008-Olsson} and plane parameter based linear equation solver (denoted as LinEq) \cite{ICPR-2021-Favre, ISPRS-2016-Khoshe, ICCVW-2017-forstner}. In detail, Olsson's method employed \textit{point-to-plane} registration using points identified as inliers, and LinEq obtained a least-squares solution from the plane coordinate transformation \cite{MVG-2023-Hartley}. Also, we denote our BnB solver as BnB and subsequent refinement of LinEq with our cost function by LM algorithm as Refine.

\noindent \textbf{Evaluation Metrics:} Estimated rotation and translation, $\hat{\mathbf{R}}$ and $\hat{\mathbf{t}}$, were evaluated by $\cos^{-1}((\mathrm{tr}(\hat{\mathbf{R}}^\top \mathbf{R}_{gt})-1)/2)$ and $\|\hat{\mathbf{t}} - \mathbf{t}_{gt}\|/ \|\mathbf{t}_{gt}\| \times 100 \%$ respectively, where $\mathbf{R}_{gt}$ and $\mathbf{t}_{gt}$ denote ground truth value.

\begin{figure}[t]
    \centering
    \includegraphics[width=1.0\columnwidth]{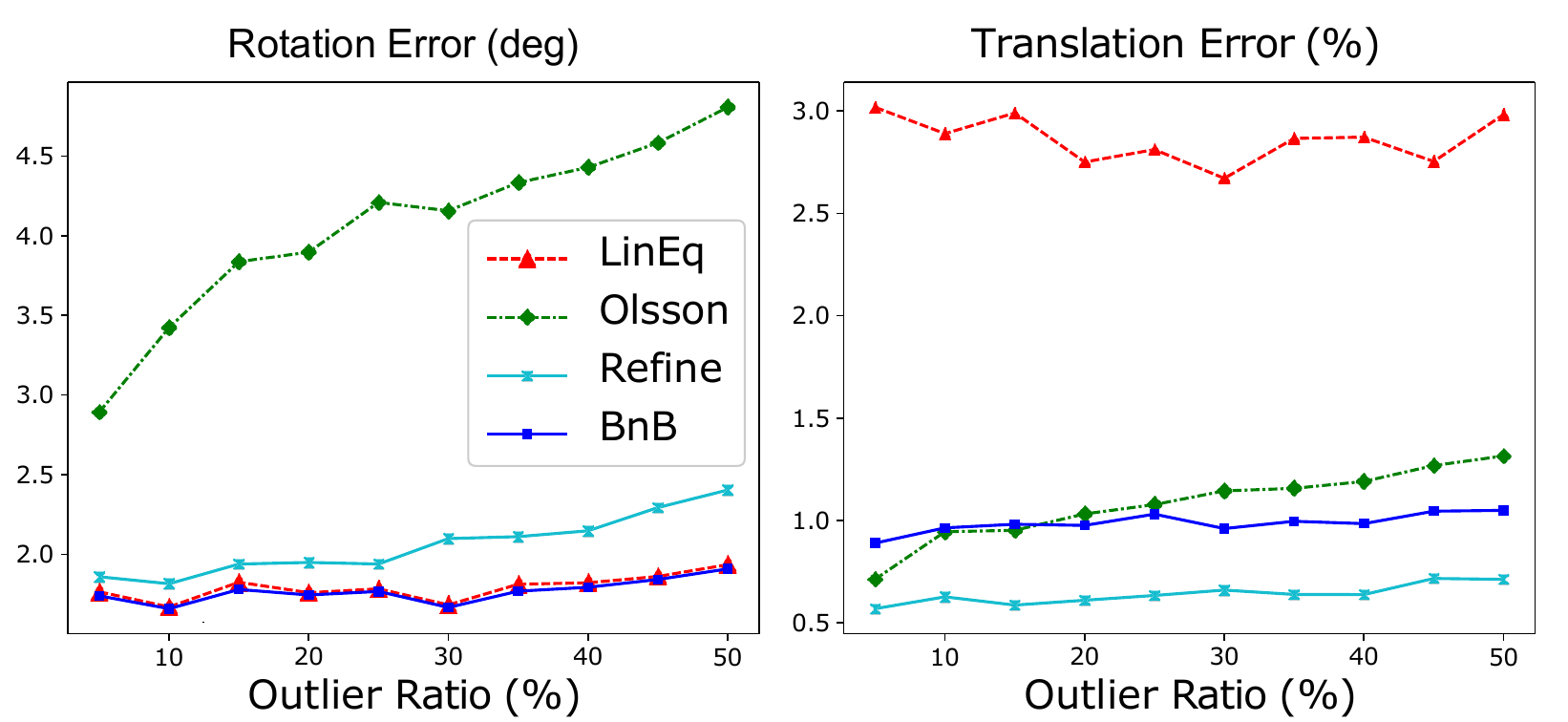}
    \caption{\textbf{Space Station data registration results.} The point-based method (Olsson) yields a severe performance drop when estimating rotation due to its vulnerability to noise. Parameter-based (LinEq) alignment fails more critically in estimating translation because explicitly selecting one of two antipodal parameters distorts the translation cost.
    }
    \label{fig:SpaceGraph}
    \vspace{-5mm}
\end{figure}

As can be seen, \figref{fig:SpaceGraph} highlights the contrast between conventional point-based (Olsson) and parameter-based (LinEq) approaches in terms of rotational and translational error. The error from Olsson's method steadily increases with respect to the outlier ratio as the plane fitting fails to remove outliers. 
The impact is particularly notable for rotational accuracy, as even 1\% of outliers that cannot be filtered out severely deteriorate the result. With such a small outlier ratio, the estimated transformation can greatly vary, especially when the noisy points move farther from the model centroid, as further demonstrated in \apref{appendix:PointVar}. 

In terms of rotational error, both LinEq and BnB showed similar performance, as they estimate rotation from the normal vector, with Refine following next. In translation error, LinEq reported the large error, as it obtains the translation vector by minimizing the difference between the last elements of the plane parameter. This constraint seeks to minimize the straight-line distance within its embedded 4-dimensional Euclidean space, often longer than the shortest path (see \apref{appendix:CurveCompare} for details). However, additional refinement of LinEq by our cost function highly decreased the translation errors, which are also lower than the result of \ac{BnB}. By trading off rotational accuracy, aligning the features through minimizing a total cost function on the manifold significantly enhances translation accuracy compared to separate estimation using BnB and LinEq.




\subsection{RGB-D Odometry}\label{Exp:RGBDodom}

Next, we demonstrate our 3D line registration for RGB-D visual odometry in indoor environments.

\noindent \textbf{Datasets:} We employed \textit{Office} and \textit{Room} sequence of Replica Dataset \cite{straub2019replica} obtained by \cite{CVPR-2022-Zhu}. 

\noindent \textbf{Baselines:} For validation, we compared three algorithms: Olsson's method, a RGB-D odometry algorithm by Park \cite{ICCV-2017-Park}, and linear solver from Pl\"{u}ckerNet \cite{CVPR-2021-Liu}. Similar to the object registration task, we selected Olsson's method as a point-based approach, performing \textit{point-to-line} registration by extracting two endpoints from each matched 3D line. Additionally, we evaluated Park's method as another point-based approach, which performs colored point cloud registration. For parameter-based approach, we used back-end solver of Pl\"{u}ckerNet, denoted as LinEq, which is a recent state-of-the-art linear solver that utilizes Plücker coordinates for parameterization. We also refined the result of LinEq by our cost, denoting the method as Refine. We extracted and matched 2D line segments from consecutive RGB images with GlueStick \cite{ICCV-2023-Pautrat} and generated corresponding 3D line pairs from depth images.

\noindent \textbf{Evaluation Metrics:} Relative poses between consecutive frames were accumulated to construct a trajectory and evaluated by \ac{APE} \cite{IROS-2012-Strum}. 

\begin{table}[t]\centering
    \caption{\textbf{Replica dataset RGB-D odometry evaluation.} The trajectories obtained from the accumulated relative poses of each algorithm were evaluated using APE, with rotation measured in degrees and translation in meters. Failure in odometry is denoted as a hyphen (\textemdash). The notation is shortened to `Off' for \textit{Office} and `Rm' for \textit{Room} sequences.}\label{tab:replica_result}
\begin{adjustbox}{width=1\linewidth}
\begin{tabular}{cl|lllll|lll}
\hline
\multicolumn{1}{l}{}        &        & Off0                    & Off1                      & Off2           & Off3           & Off4                      & Rm0            & Rm1            & Rm2            \\ \hline
\multirow{2}{*}{Olsson \cite{ICPR-2008-Olsson}}        & $\text{APE}_{t}$ & 0.34                   & 0.27                     & 0.38          & 0.89         & 2.46                    & 0.68          & \textemdash         & 0.61         \\
                            & $\text{APE}_{r}$ & 11.45                 & 10.06                   & 12.37        & 33.92        & 74.64                   & 30.52         & \textemdash        & 38.53        \\ \hline
\multirow{2}{*}{Park \cite{ICCV-2017-Park}}     & $\text{APE}_{t}$ & 0.48                   & 0.19                     & 0.89          & 1.37          & 1.24                     & 0.48          & 0.73          & 0.36          \\
                            & $\text{APE}_{r}$ & 48.93                  & 20.38                    & 33.85         & 75.32         & 60.63                    & 20.58         & 28.81         & 15.59         \\ \hline 
\multirow{2}{*}{LinEq~\cite{CVPR-2021-Liu}} & $\text{APE}_{t}$ & 0.16                   & 0.33 & 0.19          & 0.40          & 0.22 & 0.29          & 0.91          & 0.55          \\
                            & $\text{APE}_{r}$ & 8.83                   & 22.72                    & 6.38          & 7.06          & 5.98                     & 10.36         & 108.59        & 13.06         \\ \hline \hline
\multirow{2}{*}{Refine}       & $\text{APE}_{t}$ & \textbf{0.05}          & \textbf{0.04}            & 0.07 & \textbf{0.09} & \textbf{0.06}            & \textbf{0.06} & \textemdash & \textbf{0.07} \\
                            & $\text{APE}_{r}$ & 4.51 & \textbf{4.20}            & 4.62 & 3.25 & \textbf{3.93}            & 2.60 & \textemdash & \textbf{3.77} \\ \hline
\multirow{2}{*}{BnB}       & $\text{APE}_{t}$ & 0.08          & 0.11            & \textbf{0.06} & 0.11 & 0.10            & 0.08 & \textbf{0.06} & 0.09 \\
                            & $\text{APE}_{r}$ & \textbf{3.91} & 7.35            & \textbf{2.29} & \textbf{2.59} & 4.71            & \textbf{2.25} & \textbf{3.14} & 5.59 \\ \hline
\end{tabular}
\end{adjustbox}
\vspace{-5mm}
\end{table}

\begin{figure}[t!]
\centering\includegraphics[width=1.0\columnwidth]{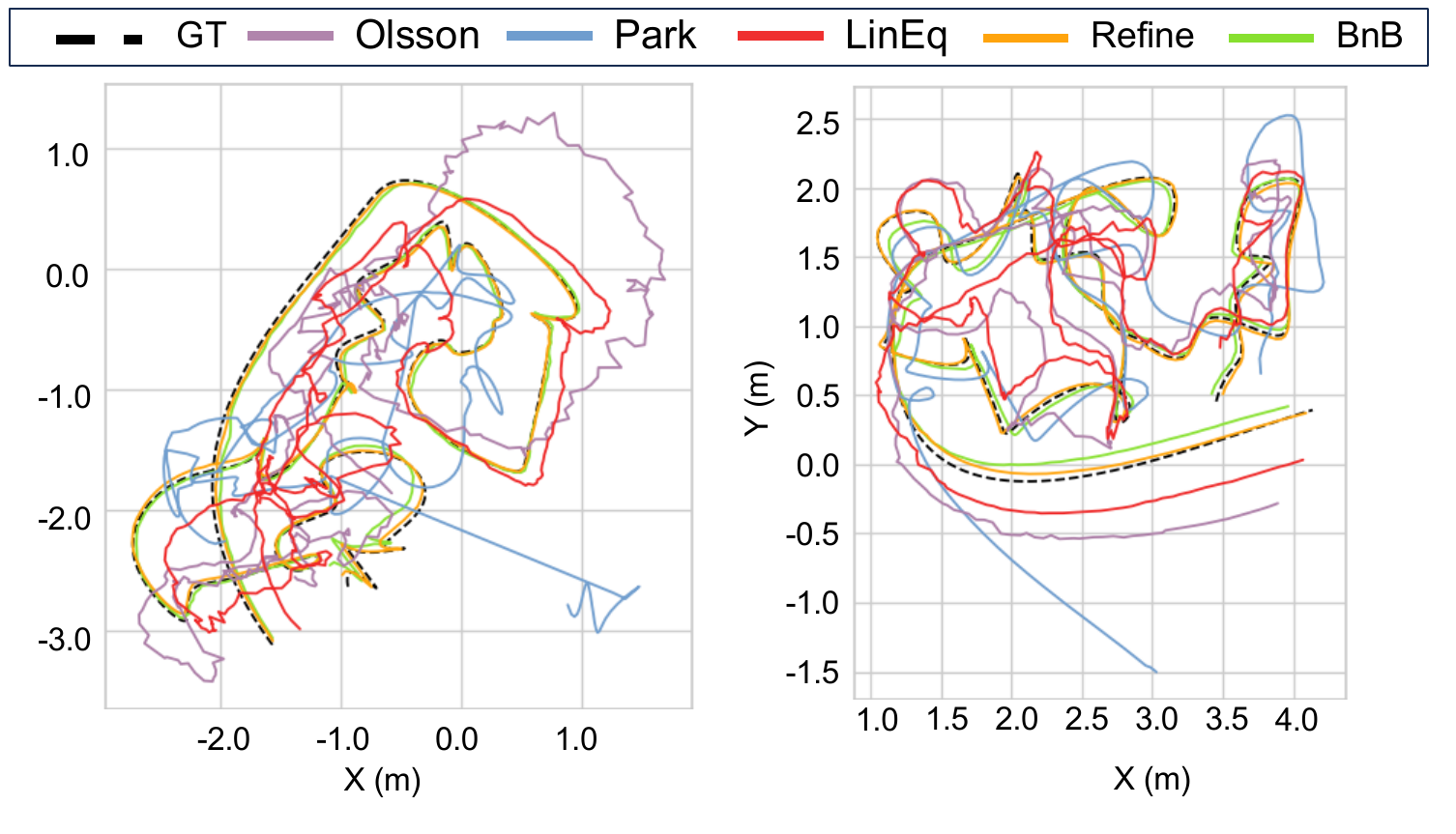}
    \caption{\textbf{Estimated trajectories of Replica dataset.} For both \textit{Office3} (left) and \textit{Room0} (right), all other benchmarks exhibit significant misalignment with the ground truth trajectory, in contrast to the high accuracy achieved by our approaches. 
     }
    \label{fig:replica_traj}
    \vspace{-5mm}
\end{figure}

The results are presented in \tabref{tab:replica_result} and \figref{fig:replica_traj}. Using 3D lines paired in RGB images, our algorithms produced the best results in all sequences.
The \textit{Room1} sequence is particularly challenging for all algorithms.
In the \textit{Room1} sequence, Olsson’s method and Refine failed, and LinEq exhibited its highest error across all sequences.
The significant error in this sequence is primarily due to frequent mismatches among lines caused by repetitive wall patterns. Failure for filtering outliers in this case by LinEq also leads to divergence of optimization result by Refine.
While Park's method could handle this scenario, it exhibited larger errors in most sequences. For instance, in the \textit{Office3} sequence, a substantial error occurred due to a single-colored wall occupying nearly half of the image.
Park's method computes the color gradient by solving a least-squares optimization of the color function; however, redundant pixels with identical intensity disrupt the gradient computation, resulting in severe trajectory drift. Detailed illustration of each failure case is provided in \apref{appendix:OdomFail}.



\subsection{Perspective-n-Line}\label{Exp:PnL}

Following the literature \cite{ICRA-2011-Mirzaei, PAMI-2016-Xu,RAL-2020-Liu}, we reformulated the \ac{PnL} problem into \textit{line-to-plane} registration by back-projecting 2D line into the 3D plane containing the origin. 

\noindent \textbf{Datasets:} We used both synthetic and real data for the validation. For simulation, we generated the synthetic data by scattering 2D end-points within the range of $[0,640] \times [0,480]$ as in \cite{PAMI-2016-Xu}. We also divide the sets into centered and uncentered cases, with the centered case having evenly distributed lines across the entire image, and the uncentered case having lines only on one quarter of the image. For both cases, we back-projected them with random depth to construct paired 3D lines. In uncentered case, 2D line segments are not uniformly distributed within the image, providing a challenging scenario prone to local minima. Increasing the outlier ratio to 80\%, we randomly generated 500 sets for a single ratio, each set containing 100 pairs. 

For the real image sequence test, chessboard images captured by a camera and motion capture system were used to evaluate the PnL results. To generate line pairs, we extracted vertical and parallel 3D lines, along with their corresponding 2D lines in the images, based on the known configuration of the chessboard.

\noindent \textbf{Baselines:} We employed two conventional PnL methods in image space, MinPnL \cite{RAL-2020-Zhou} and CvxPnL \cite{JMIV-2023-Agostinho}. Both methods represent a 3D line using two endpoints, which are then projected onto the image to construct the 2D \textit{point-to-line} cost in image space. As mentioned, another approach reformulates the problem into 3D \textit{line-to-plane} registration by interpreting the 3D plane as the preimage of the 2D line. For comparison with this approach, we also evaluated ASPnL \cite{PAMI-2016-Xu}, ASP3L \cite{PAMI-2016-Xu}, and RoPnL \cite{RAL-2020-Liu}. To enhance robustness against outliers, ASP3L employs \ac{RANSAC}, while RoPnL uses inlier-set maximization with \ac{BnB} for obtaining the rotation. We also denote Refine as refined result of ASP3L, which is only considered in chessboard pose estimation since high outlier ratio of synthetic data leads to divergence in optimized result.
\begin{figure}[t!]
    \centering
\includegraphics[width=1.0\columnwidth]{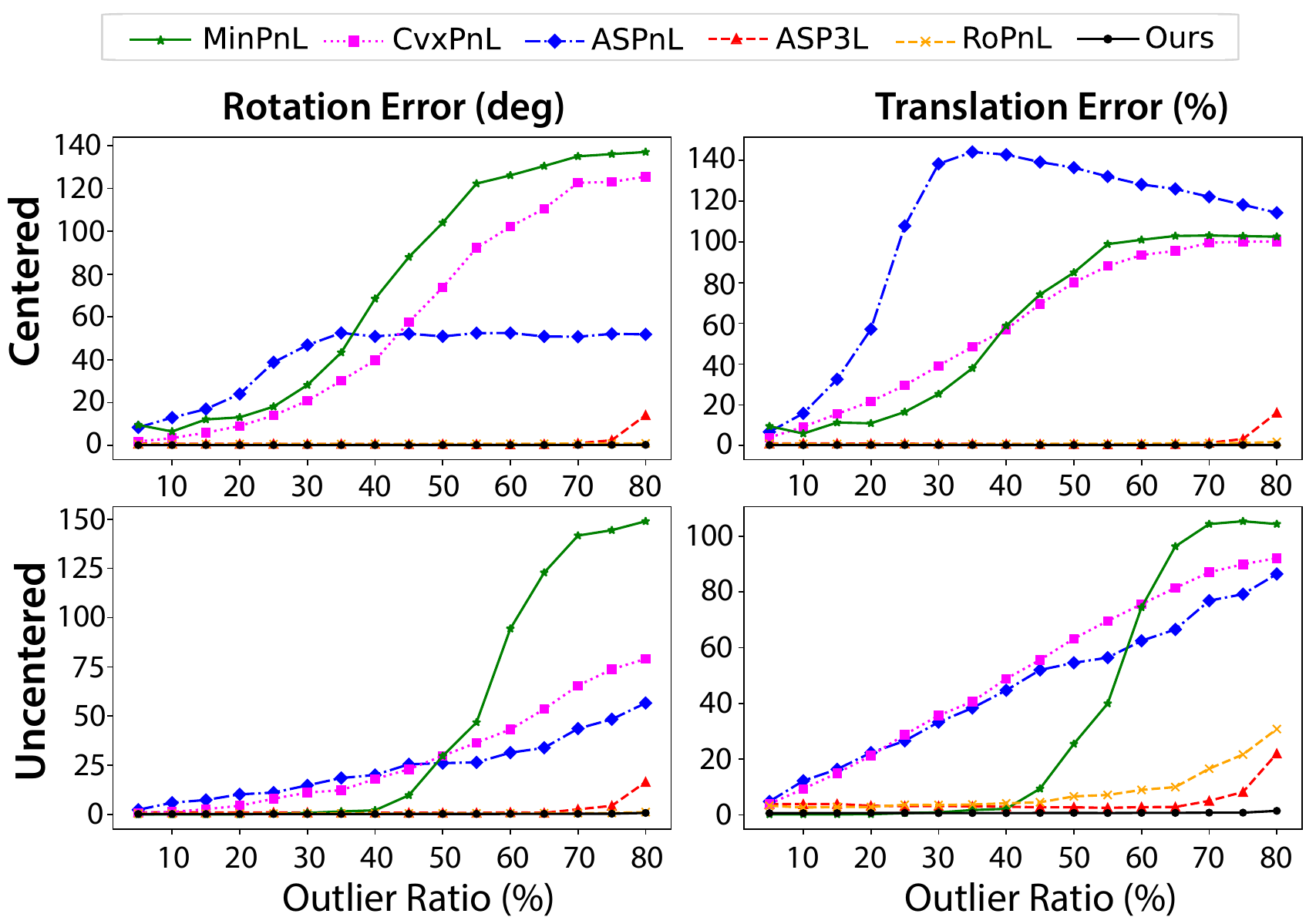}
    \caption{\textbf{PnL experiment results with synthetic data for centered (top) and uncentered (bottom) cases.} Our algorithm robustly estimates rotation and translation by solving the inlier set maximization problem, even under the outlier ratio of 80\%. 
    }
    \vspace{-5mm}
    \label{fig:Graph_PnL}
\end{figure} 

\begin{figure}[t!]
\centering\includegraphics[width=1.0\columnwidth]{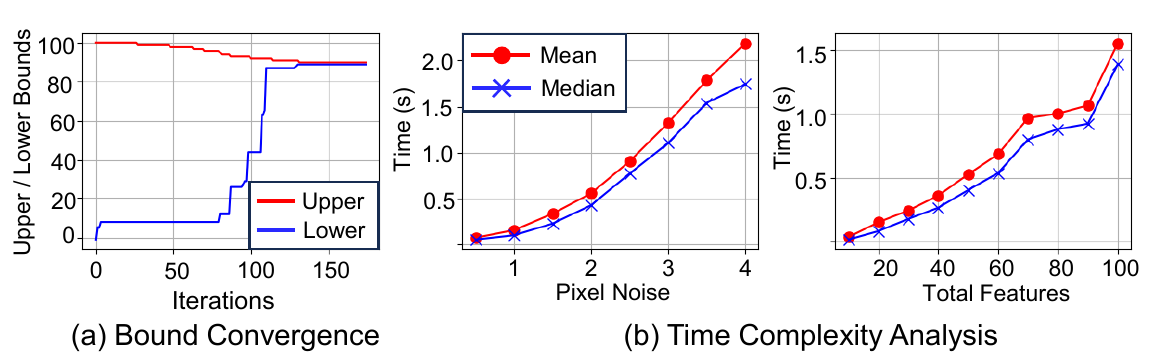}
    \vspace{-5mm}
\caption{\textbf{Additional results from the synthetic PnL experiment.} {(a) Convergence of the bounds on the inliers for 100 correspondences with a 10\% outlier ratio.} (b) Runtime with respect to endpoint pixel noise and the number of correspondences.}
    \label{fig:Bound}
    \vspace{-5mm}
\end{figure}

\noindent \textbf{Evaluation Metrics:} For the synthetic dataset, we used the same evaluation metrics for the rotation and translation error defined in \cref{Exp:Space}. For the real-image experiment, we evaluate the 6D relative pose between the camera and the chessboard in terms of rotation and translation.


Tested with synthetic data as in \figref{fig:Graph_PnL}, our registration approach consistently achieved the lowest error across all scenarios, even under a severe outlier ratio of 80\%---a level at which every other algorithms failed.
Most algorithms exhibited increasing errors as the outlier ratio rose, except for ASP3L and RoPnL, owing to the robustness achieved through outlier handling. 
Even with these two robust methods, notable translation errors were observed as the outlier ratio increased. As shown in previous experiments, our formulation stands out in handling translational costs. Specifically, although RoPnL uses the same BnB as a back-end solver for rotation, it applies a \textit{point-to-plane} distance with linear fitting on translation estimation, which led to significantly higher translation errors in the uncentered case. \figref{fig:Bound}\iccvbl{a} additionally illustrates the convergence of our rotational lower and upper bounds, empirically validating the correctness of our bound in addition to its theoretical soundness. Furthermore, we analyze the effects of line endpoint pixel noise and the number of correspondences in our BnB algorithm, as shown in \figref{fig:Bound}\iccvbl{b}.


The robustness of ASP3L, RoPnL, and \ac{BnB} in filtering out noisy correspondences was also evident in real images, resulting in lower errors compared to other methods as listed in \tabref{tab:PnL_result}. Specifically, our approaches achieved the best results by providing a deterministic optimum for \textit{line-to-plane} registration, outperforming RoPnL and ASP3L.

\begin{figure}[t!]
\centering\includegraphics[width=1.0\columnwidth]{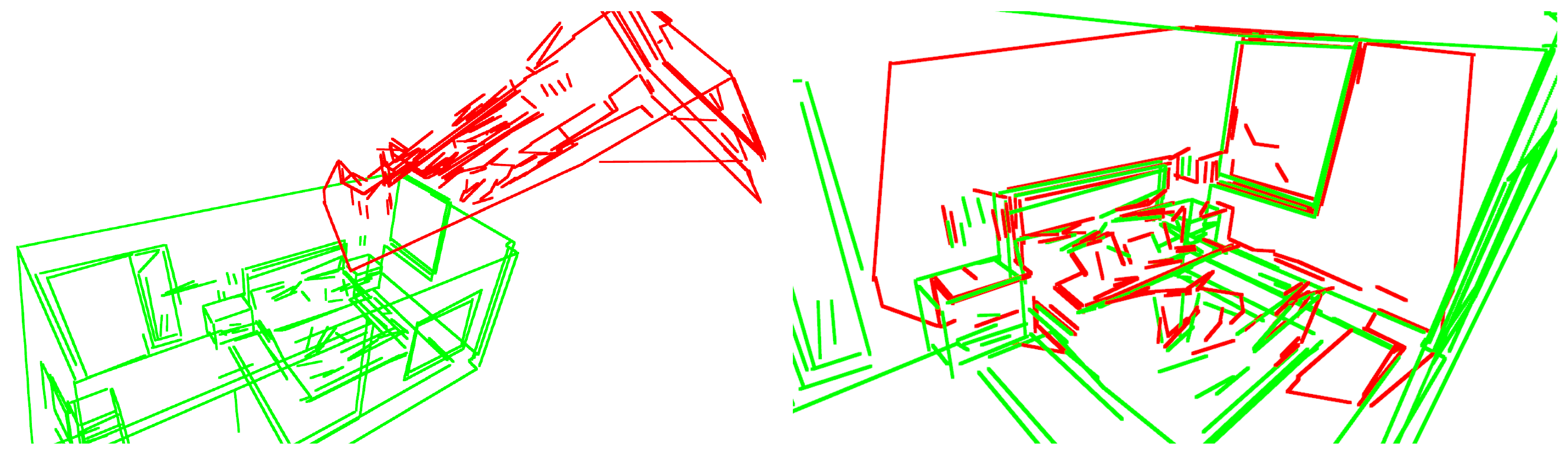}
    \caption{Our work can also align partially overlapped line reconstruction without correspondences. \textbf{Left}: \textcolor{green}{Green} and \textcolor{red}{Red} lines are extracted from the mesh and a depth image of \textit{Room1} sequence of Replica dataset \cite{straub2019replica}. \textbf{Right}: Two line reconstructions are successfully aligned using our BnB extension for correspondence search.}
    \label{fig:rebuttal}
    \vspace{-3mm}
\end{figure}
\begin{table}[t]
    \caption{\textbf{Experiment results for 6D chessboard pose estimation.} Our method outperformed all other \ac{PnL} algorithms in both parameters, demonstrating its effectiveness with real-world data. }\label{tab:PnL_result}
\centering
\begin{adjustbox}{width=1\linewidth}
\begin{tabular}{l|c|c|c|c|c|c|c}
\hline  
\multicolumn{1}{c|}{} & MinPnL \cite{RAL-2020-Zhou} & CvxPnL \cite{JMIV-2023-Agostinho}   & ASPnL \cite{PAMI-2016-Xu}    & ASP3L \cite{PAMI-2016-Xu}  & RoPnL \cite{RAL-2020-Liu}  & Refine & BnB   \\ \hline \hline $\Delta\mathbf{R}$ (\degree)       &  0.36      &  0.53   & 0.44    & 0.30   &  0.36 & 0.38 & \textbf{0.25}  \\
$\Delta\mathbf{t}$ (mm)       &  3.49  &    5.10  & 4.60 & 1.91 & 1.43  & \textbf{1.13} & 1.23 \\ \hline
\end{tabular}
\end{adjustbox}
\vspace{-5mm}
\end{table}

\subsection{Correspondence-free Localization}\label{Exp:Localization}
In this experiment, we show our work can be extended to the correspondence-free registration problem through RGB-D image localization task. Our goal is to register a frame represented by 3D lines to the 3D line map extracted from the mesh as \figref{fig:rebuttal}. An algorithm for this task can be readily implemented by extending our \ac{BnB} solver to also search for correspondences. As a result, our algorithm successfully localized the frame within the scene, achieving an accuracy of 0.65\degree and 0.03\% error without any prior information of correspondences. Note that this task has only been possible in point cloud \cite{PAMI-2015-Yang}, and is now made solvable through our novel cost functions and \ac{BnB} solver.

\section{Conclusion}
\label{sec:conclusion}

This paper proposes an optimizable registration cost for lines and planes that minimizes the Grassmann distance. Through experiments on various problems incorporating the registration task, our approach has been validated to improve the accuracy of linear solvers and has demonstrated robustness in the presence of noisy correspondences. As foundational research defining a general cost function for subspace registration, our work can be applied to various tasks that align lines and planes. One example is scalable sensor localization via affine features. While putative correspondences can be obtained with a compatibility graph, many works still rely on inaccurate cost functions based on vector parameter \cite{IROS-2022-Lusk, arxiv-2024-Yu}. Our cost function is expected to improve localization results in such cases. Additionally, our BnB solver may serve as a frontend algorithm for line and plane-based SLAM due to its ability to find a deterministically optimal solution in challenging scenarios.

\vspace{5mm}
\noindent \textbf{Acknowledgments.} This work was supported by the NRF (No. RS-2024-00461409, No. RS-2023-00241758).

\newpage 

{
    \small
    \bibliographystyle{ieeenat_fullname}
    \bibliography{string-short,main}
}


\clearpage
\setcounter{page}{1}
\onecolumn
\begin{center}
    \Large
    \textbf{Registration beyond Points: General Affine Subspace \\ Alignment via Geodesic Distance on Grassmann Manifold}    
    \vspace{0.4cm}
    \\
    Supplementary Material
\end{center}
\section{Proof of Theorems}
\subsection{Proof of Theorem 1} \label{Appendix:Theorem1}
\begin{proof}
We need to show that $f$ satisfies the following two properties of the group action.
\begin{align}
   &\text{1}. {}^\forall \mathbf{X}\in \mathrm{Gr}(k,n), \; \mathbf{I} \cdot \mathbf{X} = \mathbf{X}\; (Identity) \nonumber \\ 
   &\text{2}. {}^\forall (\mathbf{X}\in\mathrm{Gr}(k,n),\; \mathbf{T}_1,\mathbf{T}_2 \in SE(n)),\; (\mathbf{T}_1\mathbf{T}_2) \cdot \mathbf{X} = \mathbf{T}_1 \cdot (\mathbf{T}_2 \cdot \mathbf{X}) \; (Compatibility)  \nonumber
\end{align}
1. For the identity matrix $\mathbf{I}$, it is straightforward that $\mathbf{I}\cdot\mathbf{X} = \mathbf{X}$, as the rotation matrix is the identity, and the translation is the zero vector: 
\begin{align}
    \mathbf{I}\cdot (\mathbb{A}+\mathbf{b}) &= (\mathbf{I}\cdot \mathbb{A}) + (\mathbf{Ib} + \mathbf{I}(\mathbf{I} - \mathbf{AA}^\top)\mathbf{I}^\top\mathbf{0}) \\
    &= \mathbb{A} + \mathbf{b} \nonumber
\end{align}
2. Given two elements of $SE(n)$, $\mathbf{T}_1=(\mathbf{R}_1,\mathbf{t}_1)$ and $\mathbf{T}_2=(\mathbf{R}_2,\mathbf{t}_2)$, $\mathbf{T}_1\cdot(\mathbf{T}_2\cdot\mathbf{X})$ is derived by following process:
\begin{align}
    \mathbf{T}_2 \cdot \mathbf{X} &= (\mathbf{R}_2\cdot\mathbb{A}) + \mathbf{R}_2\mathbf{b} + \mathbf{R}_2(\mathbf{I} - \mathbf{AA}^\top)\mathbf{R}_2^\top\mathbf{t}_2 \\ 
    \mathbf{T}_1 \cdot (\mathbf{T}_2 \cdot \mathbf{X}) &= \mathbf{R}_1 \cdot (\mathbf{R}_2 \cdot \mathbb{A}) + \mathbf{R}_1(\mathbf{R}_2\mathbf{b} +  \mathbf{R}_2(\mathbf{I} - \mathbf{AA}^\top)\mathbf{R}_2^\top\mathbf{t}_2) +\mathbf{R}_1(\mathbf{I} - \mathbf{R_2A}(\mathbf{R_2A})^\top)\mathbf{R}_1^\top\mathbf{t}_1 \\
    &=  \mathbf{R}_1 \cdot (\mathbf{R}_2 \cdot \mathbb{A}) + \mathbf{R}_1\mathbf{R}_2(\mathbf{b} + (\mathbf{I} - \mathbf{AA}^\top)(\mathbf{R}_2^\top\mathbf{t}_2 + \mathbf{R}_2^\top\mathbf{R}_1^\top\mathbf{t}_1)) \; (\because \mathbf{I} = \mathbf{R}_2\mathbf{R}_2^\top). \\
    &= ((\mathbf{R}_1\mathbf{R}_2) \cdot \mathbb{A}) + \mathbf{R}_1\mathbf{R}_2\mathbf{b} + \mathbf{R}_1\mathbf{R}_2(\mathbf{I}-\mathbf{AA}^\top)(\mathbf{R}_1\mathbf{R}_2)^\top(\mathbf{R}_1\mathbf{t}_2+\mathbf{t}_1) \; (\because (\mathbf{R}_2\cdot\mathbb{A}) := \mathrm{span}\{\mathbf{R}_2\mathbf{A}\}) \label{Equ:Thm1}
\end{align}
Also, since $\mathbf{T}_1\mathbf{T}_2 = (\mathbf{R}_1\mathbf{R}_2, \mathbf{R}_1\mathbf{t}_2+\mathbf{t}_1)$, 

\begin{equation}
    (\mathbf{T}_2\mathbf{T}_1)\cdot \mathbf{X} = (\mathbf{R_1}\mathbf{R}_2)\cdot \mathbb{A} + \mathbf{R}_1\mathbf{R}_2\mathbf{b} + \mathbf{R}_1\mathbf{R}_2(\mathbf{I} - \mathbf{AA}^\top)(\mathbf{R}_1\mathbf{R}_2)^\top(\mathbf{R}_1\mathbf{t}_2+\mathbf{t}_1),
\end{equation}
which is identical to \equref{Equ:Thm1}.

\end{proof}

\subsection{Proof of Corollary 1.1} \label{Appendix:Corollary1.1}
\begin{proof}
    Assume $\mathbf{x} \in \mathbb{R}^n$ is included in affine subspace $\mathbb{A}+\mathbf{b}$. Then, there exists a unique unit vector $\mathbf{c}$ satisfying 
    \begin{equation}
        \mathbf{x} = \mathbf{Ac} + \mathbf{b},
    \end{equation}
which represents a coordinate of the point on the plane. Then, by $SE(n)$ transformation, $\mathbf{x}$ moves to 
\begin{equation}\label{equ:Cor1.1_1}
    \mathbf{x}'=\mathbf{RAc} +\mathbf{Rb} + \mathbf{t}.
\end{equation}
To demonstrate that $\mathbf{x}'$ is included in $\mathbf{T}\cdot(\mathbb{A} + \mathbf{b})$, we need to show that the projection of the difference between $\mathbf{x}'$ and the displacement of $\mathbf{T}\cdot(\mathbb{A} + \mathbf{b})$ onto the orthogonal complement of $\mathbf{R}\cdot\mathbb{A}$ results in the zero vector. This can be proved as follows:

\begin{align}
    (\mathbf{I} - \mathbf{RAA}^\top\mathbf{R}^\top)(\mathbf{x}'-\mathbf{b}'(\mathbf{R},\mathbf{t})) &= (\mathbf{I} - \mathbf{RAA}^\top\mathbf{R}^\top)(\mathbf{RAc} + \mathbf{Rb} + \mathbf{t} - \mathbf{Rb} - \mathbf{R}(\mathbf{I} - \mathbf{A}\mathbf{A}^\top)\mathbf{R}^\top\mathbf{t}))\\ &= (\mathbf{I} -\mathbf{RAA}^\top\mathbf{R}^\top)\mathbf{RA}(\mathbf{c} + \mathbf{A}^\top\mathbf{R}^\top \mathbf{t}) \\ &= \mathbf{0}. \nonumber
\end{align}
\end{proof}
\noindent 

\subsection{Proof of Theorem 2} \label{Appendix:Theorem2}
\begin{proof}

($\rightarrow$) Given two elements of the Grassmannian, $\mathbb{A} \in \mathrm{Gr}(k,n)$ and $\mathbb{B} \in \mathrm{Gr}(l,n)$, where $k \leq l < n$ are positive integers, assume that every basis vector of $\mathbb{A}$, denoted as $\mathbf{a}_i$ $(i=1,\cdots,k)$, is spanned by the basis vectors of $\mathbb{B}$, denoted as $\mathbf{b}_i$ $(i=1,\cdots,l)$. In this case, each $\mathbf{a}_i$ can be represented as:
\begin{equation}
\mathbf{a}_i = \sum^l_{j=1} c_{ij}\mathbf{b}_j  
\end{equation}

\noindent This leads to orthonormal basis representation of $\mathbb{A}$ as:
\begin{equation}
    \mathbf{A} = \begin{bmatrix}
\sum^l_{j=1}c_{1j}\mathbf{b}_j & \cdots & \sum^l_{j=1}c_{kj}\mathbf{b}_j \\
\end{bmatrix} = \begin{bmatrix}
c_{11}\mathbf{b}_1 & \cdots & c_{k1}\mathbf{b}_1 \\
\end{bmatrix} + \cdots + \begin{bmatrix}
c_{1l} \mathbf{b}_l& \cdots & c_{kl}\mathbf{b}_l \\
\end{bmatrix}.
\end{equation}
By multiplying the orthonormal basis matrix of $\mathbb{B}$, which is $\mathbf{B}=\begin{bmatrix}
    \mathbf{b}_1 & \cdots & \mathbf{b}_l \\
\end{bmatrix}$, to the $\mathbf{A}^\top$, we obtain following matrix $\mathbf{A}^\top\mathbf{B} \in \mathbb{R}^{k\times l}$:

\begin{equation}
    \mathbf{A}^\top\mathbf{B} = \begin{bmatrix}
c_{11} & \dots & c_{1l} \\
\vdots  & \ddots & \vdots \\
c_{k1} & \dots & c_{kl} \\
\end{bmatrix}.
\end{equation}
Since every $\mathbf{a}_i$ is an orthonormal matrix, we have the following two conditions:
\begin{align}
     &c_{i1}^2 +  c_{i2}^2 + \dots + c_{il}^2 = 1, \; (i=1,\dots,k) \\ 
     &\sum^l_{j=1} c_{ij}c_{i'j} = 0 \; (\forall i, i' \in \{1,\dots,k\},\; i \neq i').
\end{align} This implies that $\mathbf{A}^\top\mathbf{B}$ consists of $k$ orthonormal row vectors. Since every singular value of an orthonormal matrix is 1, all principal angles are equal to zero by \equref{equ:prinAng}, resulting in a Grassmann distance of zero.
\\

\noindent ($\leftarrow$) Given two elements of the Grassmannian, $\mathbb{A} \in \mathrm{Gr}(k,n)$ and $\mathbb{B} \in \mathrm{Gr}(l,n)$, where $k \leq l < n$ are positive integers, assume that the Grassmann distance between them is zero. Then, for every principal vector pair $(\mathbf{p}_i, \mathbf{q}_i)$ $(i=1, \dots, k)$, the condition $\mathbf{p}_i = \mathbf{q}_i$ is satisfied. According to \defref{def:GrassmannDist}, these principal vectors constitute the first $k$ orthonormal basis vectors of both $\mathbb{A}$ and $\mathbb{B}$. Therefore, every basis vector of $\mathbb{A}$ is spanned by the basis vectors of $\mathbb{B}$.
\end{proof}

\subsection{Proof of Problem 2} \label{Appendix:Problem2}
Given two elements of the affine Grassmannian, $\mathbb{A} + \mathbf{c} \in \mathrm{Graff}(k,n)$ and $\mathbb{B} + \mathbf{d} \in \mathrm{Graff}(l,n)$, where $k \leq l < n$ are positive integers, we need to show that every basis vector of $z(\mathbb{A} + \mathbf{c})$ is spanned by the basis vectors of $z(\mathbf{T} \cdot (\mathbb{B} + \mathbf{d}))$ if and only if the following condition is satisfied:

\begin{equation}\label{Appendix:BasisSpan}
    \sum_{i=1}^{k}\left\|\mathbf{P}_{\mathbf{R} \cdot \mathbb{B}}\mathbf{a}_i-\mathbf{a}_i\right\|_2^2 +
\left\|\mathbf{P}_{z(\mathbf{T} \cdot (\mathbb{B}+\mathbf{d}))}\tilde{\mathbf{c}}- \tilde{\mathbf{c}}\right\|_2^2 = 0.
\end{equation}

\begin{proof}
($\rightarrow$) The orthonormal basis matrix of each embedded subspace is represented as:
\begin{equation}
    \mathbf{Y}_{z(\mathbb{A} + \mathbf{c})} = \begin{bmatrix}
\mathbf{a}_1 & \dots &\mathbf{a}_k  & \frac{\mathbf{c}}{\sqrt{1+\left\| \mathbf{c}\right\|^2}} \\
 0& \dots &  0& \frac{1}{\sqrt{1+\left\| \mathbf{c}\right\|^2}}  \\
\end{bmatrix},   \; \mathbf{Y}_{z(\mathbf{T} \cdot (\mathbb{B} + \mathbf{d}))} = \begin{bmatrix}
\mathbf{Rb}_1 & \dots &\mathbf{Rb}_l  & \frac{\mathbf{d'(\mathbf{R},\mathbf{t})}}{\sqrt{1+\left\| \mathbf{d'(\mathbf{R},\mathbf{t})}\right\|^2}} \\
 0& \dots &  0& \frac{1}{\sqrt{1+\left\| \mathbf{d'(\mathbf{R},\mathbf{t})}\right\|^2}}  \\
\end{bmatrix}
\end{equation}

Since $\bar{\mathbf{a}}_i=[\mathbf{a}_i^\top \; 0]^\top$ is spanned by the columns of $\mathbf{Y}_{z(\mathbf{T}\cdot(\mathbb{B}+\mathbf{d}))}$, we can write:
\begin{equation}
\bar{\mathbf{a}}_i = \sum^l_{j=1} c_{ij}\overline{\mathbf{Rb}}_j + c_{i (l+1)}\tilde{\mathbf{d}}'(\mathbf{R},\mathbf{t}).
\end{equation}
It is readily shown that $c_{i(l+1)}=0$ since the last element of $\bar{\mathbf{a}}_i$ is zero. Then, $\mathbf{a}_i$ is a linear combination of $\mathbf{B} = [\mathbf{b}_1 \dots \mathbf{b}_l]$, where $c_{ij}=\mathbf{a}_i^\top\mathbf{Rb}_j$ due to the orthonormality of $\mathbf{RB}$. Rewrite this with linear combination of $\tilde{\mathbf{c}}$:
\begin{align}\label{equ:Appendix7.5_3}
&\mathbf{a}_i = \sum^l_{j=1} (\mathbf{a}_i^\top \mathbf{Rb}_j)\mathbf{Rb}_j =\mathbf{P}_{\mathbf{R}\cdot\mathbb{B}}\mathbf{a}_i,\\
&\label{equ:Appendix7.5_4}\tilde{\mathbf{c}} = \sum^l_{j=1} (\tilde{\mathbf{c}}^\top \overline{\mathbf{Rb}}_j)\overline{\mathbf{Rb}}_j + (\tilde{\mathbf{c}}^\top\tilde{\mathbf{d}'}(\mathbf{R},\mathbf{t}))\tilde{\mathbf{d}'}(\mathbf{R},\mathbf{t}) = \mathbf{P}_{z(\mathbf{T}\cdot(\mathbb{B}_\mathbf{d}))}\tilde{\mathbf{c}},
\end{align}
which is identical to the condition \equref{Appendix:BasisSpan} being satisfied.
\\

($\leftarrow$) Starting from \equref{Appendix:BasisSpan} being satisfied, we also derive \equref{equ:Appendix7.5_3} and \equref{equ:Appendix7.5_4}, which implies that every basis of $z(\mathbb{A}+\mathbf{c})$ is spanned by the bases of $z(\mathbf{T} \cdot (\mathbb{B}+\mathbf{d}))$.
\end{proof} As a result, to validate whether the bases of another affine subspace span the basis of one affine subspace, we only need to check if \equref{Appendix:BasisSpan} is satisfied. To formulate the cost function for minimizing the geodesic distance, this condition can be utilized by defining the left-hand side as a residual. Summing these residuals over $N$ correspondences yields the cost function for Problem \ref{problem:redefine}.
\subsection{Proof of Corollaries} \label{Appendix:Cor2}
\hspace*{1em} Simply replacing each affine primitive in Problem \ref{problem:redefine} with 3D lines and planes, as represented in each corollary, provides the result. 


\section{Derivation Details} 
\label{Appendix:Details}
\subsection{Grassmann Distance}\label{Appendix:GrassDist}
Inducing the geodesic distance on the Grassmannian requires a principal vector and angle defined as follows \cite{SIAM-2016-Ye}: 
\begin{definition}[Principal Vector]
\label{def:PrincipalVec}
Let $\mathbb{A} \in \mathrm{Gr}(k,n)$, $\mathbb{B} \in \mathrm{Gr}(l,n)$, and $k \leqslant l < n$ be positive integers. Then $i^\text{th}$ principal vectors $(\mathbf{p}_i,\mathbf{q}_i), i=1, \cdots, k$, are defined recursively as solutions to the optimization problem:
\begin{align}\begin{split}
&\max{(\mathbf{p}_i^\top \mathbf{q}_i)} \;\text{  subject to} \\ 
&\mathbf{p}_i\in\mathbb{A}, \mathbf{p}_i^\top \mathbf{p}_1 = \cdots = \mathbf{p}_i^\top \mathbf{p}_{i-1}=0, \left\|\mathbf{p}_i \right\|_2=1, \\
&\mathbf{q}_i\in\mathbb{B}, \mathbf{q}_i^\top \mathbf{q}_1 = \cdots = \mathbf{q}_i^\top \mathbf{q}_{i-1}=0, \left\|\mathbf{q}_i \right\|_2=1. 
\end{split}\end{align}
Then, the $i^\text{th}$ principal angle $\theta_i$ is defined by: 
\begin{equation}
    \cos{\theta_i} = \mathbf{p}^\top _i\mathbf{q}_i.
\end{equation}
\end{definition}
Principal angles provide a natural measure for obtaining the closeness between two linear subspaces within $\mathbb{R}^n$, spanned by columns of matrices \cite{golub1995canonical,hotelling1992relations}. This is due to its recursive definition, which extends the distance between 1-dimensional linear subspaces—explicitly derived as cosine similarity—to the range spaces of matrices. 
Derivation of principal angles requires \ac{SVD} \cite{bjorck1973numerical}. Let $\mathbf{A}$ and $\mathbf{B}$ be two orthonormal basis matrices of $\mathbb{A}\in \mathrm{Gr}(k,n)$ and $\mathbb{B}\in \mathrm{Gr}(l,n)$. Then, the principal angles are given by: 
\begin{equation}\label{equ:prinAng}
    \theta_i = \cos^{-1}{\sigma_i}, \; i=1,\cdots,k,
\end{equation}
where $\sigma_i$ refers to $i^\text{th}$ singular value of $\mathbf{A}^{\top}\mathbf{B}$. Then, the geodesic distance on two elements of Grassmannian, denoted as Grassmann distance is defined as follows: 
\begin{definition}[Grassmann Distance]
\label{def:GrassmannDist}
Let  $k \leqslant l < n$, and $\theta_1, \cdots, \theta_k$ be the principal angles between $\mathbb{A} \in \mathrm{Gr}(k,n)$ and $\mathbb{B} \in \mathrm{Gr}(l,n)$, then the geodesic distance between $\mathbb{A}$ and $\mathbb{B}$ is given by: 
\begin{equation}\label{equ:GrassmannDist}
    \mathrm{d}_{\mathrm{Gr}}(\mathbb{A},\mathbb{B}) = (\sum_{i=1}^{k}\theta^2_i)^{1/2}.
\end{equation}
\end{definition} 

\subsection{Displacement Vectors} \label{Appendix:Disp}
Given an affine subspace $\mathbb{A} + \mathbf{c} \in \mathrm{Graff}(k,n)$, the unique displacement $\mathbf{c}_0$ of this space is determined as follows:
\begin{equation}
    \mathbf{c}_0 = (\mathbf{I} - \mathbf{AA}^\top)\mathbf{c}.
\end{equation}
This also represents the displacement and is orthogonal to $\mathbb{A}$ since:
\begin{align}
    \mathbf{A}^\top\mathbf{c}_0 &= \mathbf{A}^\top(\mathbf{I} - \mathbf{AA}^\top)\mathbf{c} = 0, \\ 
    (\mathbf{I} - \mathbf{AA}^\top)(\mathbf{c} - \mathbf{c}_0) &= (\mathbf{I} - \mathbf{AA}^\top)(\mathbf{c} - (\mathbf{I} - \mathbf{AA}^\top)\mathbf{c})\\  &= (\mathbf{I} - \mathbf{AA}^\top)(\mathbf{AA}^\top\mathbf{c}) \nonumber \\ &=  0 \nonumber,    
\end{align} where each equation demonstrates the orthogonality of $\mathbf{c}_0$ to $\mathbb{A}$ and the orthogonality of $\mathbf{c} - \mathbf{c}_0$ to the orthogonal complement of $\mathbb{A}$.

\subsection{Rotation Search}
\label{Appendix:RotBnB}
\subsubsection{Line-to-line case}\label{Appendix:l2l_bound}
Assume the current rotation cube is $\mathcal{C}_\mathbf{r}$ with a half-side length of $\sigma_r$ and centered at $\mathbf{R}_0$. An objective function for this case is:
\begin{equation}\label{equ:l2l_cost}
  \max_{\mathbf{R}\in \mathcal{C}_r}\sum_{i=1}^{N}\mathbf{1}(\epsilon - \mathrm{d}_\mathrm{Gr}(\mathbf{Rd}^i_1,\mathbf{d}^i_2)^2).
\end{equation}
We first derive a lower bound for $\mathrm{d}_\mathrm{Gr}(\mathbf{R}\mathbf{d}^i_1, \mathbf{d}^i_2)$ for an arbitrary rotation $\mathbf{R}$ within the cube $\mathcal{C}_\mathbf{r}$ using the triangle inequality of the Grassmann distance:
\begin{align}\label{equ:l2l_bound1}
    \mathrm{d}_\mathrm{Gr}(\mathbf{Rd}^i_1,\mathbf{d}^i_2)  \geq \mathrm{d}_\mathrm{Gr}(\mathbf{R}_0\mathbf{d}_1^i, \mathbf{d}^i_2 ) - \mathrm{d}_\mathrm{Gr}(\mathbf{R}_0\mathbf{d}^i_1,\mathbf{R}\mathbf{d}_1^i).
\end{align} Then from \cite{PAMI-2015-Yang}, an upper bound for $\mathrm{d}_\mathrm{Gr}(\mathbf{R}_0\mathbf{d}^i_1,\mathbf{R}\mathbf{d}_1^i)$ is written as:
\begin{equation}\label{equ:l2l_bound2}
    \mathrm{d}_\mathrm{Gr}(\mathbf{R}_0\mathbf{d}^i_1,\mathbf{R}\mathbf{d}_1^i) \leq \min (\frac{\pi}{2} , \sqrt{3}\sigma_r).
\end{equation}
From \equref{equ:l2l_bound1} and \equref{equ:l2l_bound2}, an upper bound for the objective function of \equref{equ:l2l_cost} is derived as:

\begin{align} \label{equ:L2LUB}\max_{\mathbf{R}\in\mathcal{C}_r}\sum_{i=1}^{N}\mathbf{1}\left(\epsilon - \mathrm{d}_\mathrm{Gr}(\mathbf{Rd}^i_1,\mathbf{d}^i_2)^2\right) &\leq  \sum_{i=1}^{N}\mathbf{1}\left(\epsilon -  \max\left(0,\mathrm{d}_\mathrm{Gr}(\mathbf{R}_0\mathbf{d}_1^i,\mathbf{d}^i_2) -\min(\frac{\pi}{2}, \sqrt{3}\sigma_r) \right)^2 \right) \\ &:= \bar{\nu}_r.
\end{align}
Additionally, a lower bound for the objective function in \equref{equ:l2l_cost} is readily derived as:
\begin{align}\label{equ:L2LLB}
    \max_{\mathbf{R}\in \mathcal{C}_r}\sum_{i=1}^{N}\mathbf{1}\left(\epsilon - \mathrm{d}_\mathrm{Gr}(\mathbf{Rd}^i_1,\mathbf{d}^i_2)^2\right) &\geq \sum_{i=1}^{N}\mathbf{1}\left(\epsilon - \mathrm{d}_\mathrm{Gr}(\mathbf{R}_0\mathbf{d}^i_1,\mathbf{d}^i_2)^2\right) \\ & := \underline{\nu}_r
\end{align}

\subsubsection{Line-to-plane case}
An objective function for this case is:

\begin{equation}\label{equ:l2p_cost}
 \max_{\mathbf{R}\in \mathcal{C}_r}\sum_{i=1}^{N}\mathbf{1}\left(\epsilon - \mathrm{d}_\mathrm{Gr}(\mathbf{d}^i, \mathbf{P}_{\mathbf{R} \cdot \mathbb{B}^i}\mathbf{d}^i)^2\right).
\end{equation} From the triangle inequality, the lower bound of $\mathrm{d}_\mathrm{Gr}(\mathbf{d}^i, \mathbf{P}_{\mathbf{R} \cdot \mathbb{B}^i}\mathbf{d}^i)$ is derived as:

\begin{equation}\label{equ:l2p_tri}
    \mathrm{d}_\mathrm{Gr}(\mathbf{d}^i, \mathbf{P}_{\mathbf{R} \cdot \mathbb{B}^i}\mathbf{d}^i) \geq \mathrm{d}_\mathrm{Gr}(\mathbf{d}^i, \mathbf{P}_{\mathbf{R}_0 \cdot \mathbb{B}^i}\mathbf{d}^i) - \mathrm{d}_\mathrm{Gr}(\mathbf{P}_{\mathbf{R}_0 \cdot \mathbb{B}^i}\mathbf{d}^i, \mathbf{P}_{\mathbf{R} \cdot \mathbb{B}^i}\mathbf{d}^i).
\end{equation} 
\noindent Denote a normal vector of $\mathbb{B}^i$ as $\mathbf{n}^i$, an acute angle between $\mathbf{R}_0\mathbf{n}^i$ and $\mathbf{Rn}^i$ as $\theta$ (it is identical to $\mathrm{d}_\mathrm{Gr}(\mathbf{R}_0\mathbf{n}^i,\mathbf{Rn}^i)$), and a mid-point of $\mathbf{R}_0\mathbf{n}^i$ and $\mathbf{R}\mathbf{n}^i$ on the arc as $\mathbf{R}_m\mathbf{n}^i$. Then, an upper bound of $\mathrm{d}_\mathrm{Gr}(\mathbf{P}_{\mathbf{R}_0 \cdot \mathbb{B}^i}\mathbf{d}^i, \mathbf{P}_{\mathbf{R} \cdot \mathbb{B}^i}\mathbf{d}^i)$ can only be explicitly derived under following observation:

\begin{equation}
    \text{if } \mathrm{d}_\mathrm{Gr}(\mathbf{R}_m\mathbf{n}^i,\mathbf{d}^i) \geq \frac{\theta}{2} \rightarrow \mathrm{d}_\mathrm{Gr}(\mathbf{P}_{\mathbf{R}_0 \cdot \mathbb{B}^i}\mathbf{d}^i, \mathbf{P}_{\mathbf{R} \cdot \mathbb{B}^i}\mathbf{d}^i) \leq \mathrm{d}_\mathrm{Gr}(\mathbf{R}_0\mathbf{n}, \mathbf{Rn}) \leq \min(\frac{\pi}{2},\sqrt{3}\sigma_r).  
\end{equation}
Extending this observation into $\mathcal{C}_r$, we obtain an upper bound $\psi_r$ within the cube:

\begin{equation}
    \psi_r = \begin{cases}
\min(\pi/2,\sqrt{3}\sigma_r) & \left(\mathrm{d}_\mathrm{Gr}(\mathbf{R}_0\mathbf{n}^i,\mathbf{d}^i) \geq \sqrt{3}\sigma_r\right) \\ 
\pi/2 & \left(\mathrm{d}_\mathrm{Gr}(\mathbf{R}_0\mathbf{n}^i,\mathbf{d}^i) < \sqrt{3}\sigma_r\right) 
\end{cases}
\end{equation} Then, the upper bound of \equref{equ:l2p_cost} is derived by substituting $\psi_r$ into \equref{equ:l2p_tri}:

\begin{align}
\max_{\mathbf{R}\in\mathcal{C}_r}\sum_{i=1}^{N}\mathbf{1}\left(\epsilon - \mathrm{d}_\mathrm{Gr}(\mathbf{d}^i,\mathbf{P}_{\mathbf{R}\cdot\mathbb{B}^i}\mathbf{d}^i)^2\right) &\leq  \sum_{i=1}^{N}\mathbf{1}\left(\epsilon -  \max\left(0,\mathrm{d}_\mathrm{Gr}(\mathbf{d}^i,\mathbf{P}_{\mathbf{R}_0\mathbb{B}^i}\mathbf{d}^i) -\psi_r \right)^2 \right) \\ &:= \bar{\nu}_r.
\end{align} A lower bound is derived similarly as \textit{line-to-line} case, which is:
\begin{align}
   \max_{\mathbf{R}\in\mathcal{C}_r}\sum_{i=1}^{N}\mathbf{1}\left(\epsilon - \mathrm{d}_\mathrm{Gr}(\mathbf{d}^i,\mathbf{P}_{\mathbf{R}\cdot\mathbb{B}^i}\mathbf{d}^i)^2\right) &\geq \sum_{i=1}^{N}\mathbf{1}\left(\epsilon - \mathrm{d}_\mathrm{Gr}(\mathbf{d}^i,\mathbf{P}_{\mathbf{R}_0\cdot\mathbb{B}^i}\mathbf{d}^i)^2\right) \\ & := \underline{\nu}_r
\end{align}

\subsubsection{Plane-to-plane case}
The same results can be obtained by substituting the direction vectors $\mathbf{d}_1^i$ and $\mathbf{d}_2^i$ with the normal vectors $\mathbf{n}_1^i$ and $\mathbf{n}_2^i$ from \cref{Appendix:l2l_bound}.

\subsection{Translation Search}
\label{Appendix:TransBnB}
\subsubsection{Line-to-line case}
Assume the current translation cube is $\mathcal{C}_\mathbf{t}$, with the center at $\mathbf{t}_0$ and vertices denoted as $\mathcal{V}_t$. An objective function for this case is:
\begin{equation}
  \min_{\mathbf{t}\in \mathcal{C}_t}\sum_{i=1}^N\left\|\mathbf{P}_{z(\mathbf{T} \cdot l^i_1)}\tilde{\mathbf{b}}^i_2-\tilde{\mathbf{b}}^i_2\right\|_2^2,
\end{equation} where $\mathbf{T} = (\mathbf{R}^*, \mathbf{t})$ and $\mathbf{P}_{z(\mathbf{T} \cdot l^i_1)}\mathbf{b}_2^i =\left((\tilde{\mathbf{b}}^i_2)^\top \overline{\mathbf{R}^*\mathbf{d}^i_1}\right)\overline{\mathbf{R}^*\mathbf{d}^i_1} + \left((\tilde{\mathbf{b}}^i_2)^\top \tilde{\mathbf{b}}^{'i}_1(\mathbf{R}^*,\mathbf{t})\right)\tilde{\mathbf{b}}^{'i}_1(\mathbf{R}^*,\mathbf{t})$. Then, from the triangle inequality of Euclidean distance, lower bound of $\left\|\mathbf{P}_{z(\mathbf{T} \cdot l^i_1)}\tilde{\mathbf{b}}^i_2-\tilde{\mathbf{b}}^i_2\right\|_2$ is: 

\begin{equation}
    \left\|\mathbf{P}_{z(\mathbf{T} \cdot l^i_1)}\tilde{\mathbf{b}}^i_2-\tilde{\mathbf{b}}^i_2\right\|_2 \geq \left\|\mathbf{P}_{z(\mathbf{T}_0 \cdot l^i_1)}\tilde{\mathbf{b}}^i_2-\tilde{\mathbf{b}}^i_2\right\|_2 - \left\|\mathbf{P}_{z(\mathbf{T}_0 \cdot l^i_1)}\tilde{\mathbf{b}}^i_2-\mathbf{P}_{z(\mathbf{T} \cdot l^i_1)}\tilde{\mathbf{b}}^i_2\right\|_2,
\end{equation} where $\mathbf{T}_0 = (\mathbf{R}^*,\mathbf{t}_0)$. From its definition, $\left\|\mathbf{P}_{z(\mathbf{T}_0 \cdot l^i_1)}\tilde{\mathbf{b}}^i_2-\mathbf{P}_{z(\mathbf{T} \cdot l^i_1)}\tilde{\mathbf{b}}^i_2\right\|_2$ can be rewritten as:

\begin{equation}\label{equ:l2l_disp_subtract}
    \left\|\mathbf{P}_{z(\mathbf{T}_0 \cdot l^i_1)}\tilde{\mathbf{b}}^i_2-\mathbf{P}_{z(\mathbf{T} \cdot l^i_1)}\tilde{\mathbf{b}}^i_2\right\|_2 = \left\|\left((\tilde{\mathbf{b}}^i_2)^\top \tilde{\mathbf{b}}^{'i}_1(\mathbf{R}^*,\mathbf{t}_0)\right)\tilde{\mathbf{b}}^{'i}_1(\mathbf{R}^*,\mathbf{t}_0) - \left((\tilde{\mathbf{b}}^i_2)^\top \tilde{\mathbf{b}}^{'i}_1(\mathbf{R}^*,\mathbf{t})\right)\tilde{\mathbf{b}}^{'i}_1(\mathbf{R}^*,\mathbf{t})\right\|_2.
\end{equation} Recall that from \thmref{equ:SE(n)Action}:

\begin{equation}
    \mathbf{b}^{'i}_1(\mathbf{R}^*,\mathbf{t}) = \mathbf{R}^*\mathbf{b}_1^i + \mathbf{R}^*(\mathbf{I}-\mathbf{dd}^\top)\mathbf{R}^{*\top}\mathbf{t}.
\end{equation} Since $\mathbf{t}\in\mathcal{C}_t$, a set of vectors $\mathbf{b}^{'i}_1(\mathbf{R}^*,\mathbf{t})$ forms a line segment within $\mathbb{R}^3$, where its two end-points are always formulated from $\mathbf{t}\in\mathcal{V}_t$. Augmenting the last element with 1 and normalizing to make $\tilde{\mathbf{b}}_1^{'i}(\mathbf{R}^*,\mathbf{t})$, the set is now mapped to an arc on 3-sphere in $\mathbb{R}^4$, and the two end-points of the arc $\mathbf{v}_1$ and $\mathbf{v}_2$ are maintained, which means $\mathbf{v}_1,\mathbf{v}_2\in\tilde{\mathbf{b}}_1^{'i}(\mathbf{R}^*,\mathbf{t}), \text{where } \mathbf{t}\in\mathcal{V}_t$. 

Our objective is to obtain an upper bound of \equref{equ:l2l_disp_subtract}. Observe that each term of the right-hand side is multiplied by $\left((\tilde{\mathbf{b}}^i_2)^\top \tilde{\mathbf{b}}^{'i}_1(\mathbf{R}^*,\mathbf{t}_0)\right)$ and $\left((\tilde{\mathbf{b}}^i_2)^\top \tilde{\mathbf{b}}^{'i}_1(\mathbf{R}^*,\mathbf{t})\right)$, respectively. These terms represent an inner product between a stationary point on 3-sphere $\tilde{\mathbf{b}}^i_2$ and points on the arc. Starting from $\mathbf{v}_1$ and heading to $\mathbf{v}_2$, we observe that in every tested case in our experiments, where the two endpoints are sufficiently close, these inner product values exhibit only four possible shapes: monotonic increasing, monotonic decreasing, convex, or concave. 
As a result, we can conclude that \equref{equ:l2l_disp_subtract} achieves its maximum at one of the vertices, and its value can be expressed as:

\begin{equation}
   \psi_t= \max_{\mathbf{t}\in\mathcal{V}_t}\left\|\left((\tilde{\mathbf{b}}^i_2)^\top \tilde{\mathbf{b}}^{'i}_1(\mathbf{R}^*,\mathbf{t}_0)\right)\tilde{\mathbf{b}}^{'i}_1(\mathbf{R}^*,\mathbf{t}_0) - \left((\tilde{\mathbf{b}}^i_2)^\top \tilde{\mathbf{b}}^{'i}_1(\mathbf{R}^*,\mathbf{t})\right)\tilde{\mathbf{b}}^{'i}_1(\mathbf{R}^*,\mathbf{t})\right\|_2 
\end{equation} Therefore, the lower bound of the objective function is:
\begin{align}\label{equ:L2L_trans_LB}
    \min_{\mathbf{t}\in \mathcal{C}_t}\sum_{i=1}^N\left\|\mathbf{P}_{z(\mathbf{T} \cdot l^i_1)}\tilde{\mathbf{b}}^i_2-\tilde{\mathbf{b}}^i_2\right\|_2^2 &\geq \sum^N_{i=1}\left(\max(0,\left\|\mathbf{P}_{z(\mathbf{T}_0 \cdot l^i_1)}\tilde{\mathbf{b}}^i_2-\tilde{\mathbf{b}}^i_2\right\|_2 - \psi_t)\right)^2 \\ &= \underline{e}_t
\end{align} Also, an upper bound is:

\begin{align}\label{equ:L2L_trans_UB}
        \min_{\mathbf{t}\in \mathcal{C}_t}\sum_{i=1}^N\left\|\mathbf{P}_{z(\mathbf{T} \cdot l^i_1)}\tilde{\mathbf{b}}^i_2-\tilde{\mathbf{b}}^i_2\right\|_2^2 &\leq\sum^N_{i=1}\left\|\mathbf{P}_{z(\mathbf{T}_0 \cdot l^i_1)}\tilde{\mathbf{b}}^i_2-\tilde{\mathbf{b}}^i_2\right\|_2^2 \\ &= \bar{e}_t
\end{align}
The process for obtaining the bounds is exactly the same for the case of the \textit{line-to-plane} and \textit{plane-to-plane} cases.
\newpage
\subsection{Algorithms}\label{sec:algorithms}
This section introduces the entire pipeline for solving the \textit{line-to-line} registration problem with our \ac{BnB} solver. The process for obtaining solutions in \textit{line-to-plane} and \textit{plane-to-plane} registration is analogous; therefore, we omit detailed algorithms for these cases.
\begin{algorithm}[h]
    \footnotesize
    \caption{Optimal 3D Line Registration}\label{Alg1}
    \hspace*{\algorithmicindent} \textbf{Input:} \\
   \hspace*{\algorithmicindent}\hspace{1em} $\mathcal{X} = \{(\textbf{d}_{t_i},\textbf{b}_{t_i})\},\; (i=1,\cdots, N)$: Target lines\\
   \hspace*{\algorithmicindent}\hspace{1em} $\mathcal{Y} = \{(\textbf{d}_{s_j},\textbf{b}_{s_j})\},\; (j=1,\cdots, M)$: Source lines\\
    \hspace*{\algorithmicindent}\hspace{1em} $\mathcal{C}_\mathbf{R}$: Initial $SO(3)$ cube for rotational BnB \\
    \hspace*{\algorithmicindent}\hspace{1em} $\mathcal{C}_\mathbf{t}$: Initial $\mathbb{R}^3$ cube for translational BnB \\
    \hspace*{\algorithmicindent}\hspace{1em} $\epsilon_\mathbf{R}$, $\epsilon_\mathbf{t}$: Threshold for rotational and translational BnB \\ 
    \hspace*{\algorithmicindent}\hspace{1em}
    $\mathcal{I}_i$: Initial correspondences
    
\hspace*{\algorithmicindent}	\textbf{Output:}\\ \hspace*{\algorithmicindent}\hspace{1em} $(\mathbf{R}^*,\mathbf{t}^*)$: Optimal transformation \\  
\hspace*{\algorithmicindent} \hspace{1em} $\mathcal{I}_f$: Resulted correspondences
    \begin{algorithmic}[1]
    \State $\mathcal{X}_{lin}$ = \{\}, $\mathcal{Y}_{lin}$ = \{\} 
    \For{$i = 1 : N$} \quad \% \defref{def:GraffCoord}
            \State $\mathcal{X}_{lin}[i][:,0] \leftarrow \bar{\mathbf{d}}_{t_i}$
            \State $\mathcal{X}_{lin}[i][:,1] \leftarrow \tilde{\mathbf{b}}_{t_i}$
    \EndFor

    \For{$j = 1 : M$} \quad \% \defref{def:GraffCoord}
            \State $\mathcal{Y}_{lin}[j][:,0] \leftarrow \bar{\mathbf{d}}_{s_j}$
            \State $\mathcal{Y}_{lin}[j][:,1] \leftarrow \tilde{\mathbf{b}}_{s_j}$
    \EndFor
    \If{$\mathcal{I}_i$ is \textbf{not} empty}
    \State
        $\mathcal{I}_f$, $\mathbf{R}^*$ = \texttt{CorrRBnB}($\mathcal{X}_{lin}$, $\mathcal{Y}_{lin}$, $\mathcal{I}_{i}$, $\mathcal{C}_\mathbf{R}$, $\epsilon_\mathbf{R}$) \quad\text{\% \cref{Alg2}}
    \Else 
        \State
        $\mathcal{I}_f$, $\mathbf{R}^*$ = \texttt{FullRBnB}($\mathcal{X}_{lin}$, $\mathcal{Y}_{lin}$,  $\mathcal{C}_\mathbf{R}$, $\epsilon_\mathbf{R}$)\quad\text{\% \cref{Alg3}}
    \EndIf
    
    \State 
    $\mathbf{t}^*$ = \texttt{TBnB}($\mathcal{X}_{lin}$, $\mathcal{Y}_{lin}$, $\mathcal{I}_f$, $\mathbf{R}^*$, $\mathcal{C}_\mathbf{t}$, $\epsilon_\mathbf{t}$) \quad\text{\% \cref{Alg4}} 
    \State \textbf{return} $\mathcal{I}_{f}$, $\mathbf{R}^*$, $\mathbf{t}^*$

    \end{algorithmic}
\end{algorithm}

\begin{algorithm}[h]
    \footnotesize
    \caption{CorrRBnB: Rotation BnB with correspondences}
    \label{Alg2}

    \hspace*{\algorithmicindent} \textbf{Input:} \\
   \hspace*{\algorithmicindent}\hspace{1em} $\mathcal{X}_{lin} = \{(\bar{\textbf{d}}_{t_i},\tilde{\textbf{b}}_{t_i})\},\; (i=1,\cdots, N)$: Target embeddings\\
   \hspace*{\algorithmicindent}\hspace{1em} $\mathcal{Y}_{lin} = \{(\bar{\textbf{d}}_{s_j},\tilde{\textbf{b}}_{s_j})\},\; (j=1,\cdots, M)$: Source embeddings\\
    \hspace*{\algorithmicindent}\hspace{1em} $\mathcal{C}_\mathbf{R}$: Initial $SO(3)$ search cube\\
    \hspace*{\algorithmicindent}\hspace{1em} $\epsilon_\mathbf{R}$: BnB threshold \\ 
    \hspace*{\algorithmicindent}\hspace{1em}
    $\mathcal{I}_i$: Initial correspondences
    
\hspace*{\algorithmicindent}	\textbf{Output:}\\ \hspace*{\algorithmicindent}\hspace{1em} $\mathbf{R}^*$: Optimal rotation \\  
\hspace*{\algorithmicindent} \hspace{1em} $\mathcal{I}_f$: Resulted correspondences
    \begin{algorithmic}[1]
    \State $\mathcal{I}_f \leftarrow \mathcal{I}_i$
    \State Add initial cube $\mathcal{C}_\mathbf{R}$ into priority queue $\mathbf{Q}_\mathbf{R}$
    \While{$\mathcal{I}_f.size < \bar{\nu}_r$}
    \State Read cube $\mathcal{C}_r$ with the greatest upper bound $\bar{\nu}_r$ from $\mathbf{Q}_\mathbf{R}$
    \For {\textbf{all} sub-cube $\mathcal{C}_{r_i}$}
    \State Compute the lower bound $\underline{\nu}_{r_i}$ \quad\text{\% \equref{equ:L2LLB}}
    \If {$\mathcal{I}_f.size < 2\underline{\nu}_{r_i}$}
    \State ($\mathcal{I}_f$, $\mathbf{R}^*$) $\leftarrow $ \texttt{LMOptimization}($\mathcal{X}_{lin}$, $\mathcal{Y}_{lin}$, $\mathcal{I}_f$, $\mathbf{R}^*$)
    \EndIf
        \State Compute the upper bound $\bar{\nu}_{r_i}$ \quad\text{\% \equref{equ:L2LUB}}
        \If{$ \mathcal{I}_f.size <\bar{\nu}_{r_i}$} 
        \State Add $\mathcal{C}_{r_i}$ to queue $\mathbf{Q}_\mathbf{R}$
        \EndIf
    \EndFor
    \EndWhile
            \State \textbf{return} $\mathcal{I}_{f}$, $\mathbf{R}^*$
    \end{algorithmic}
\end{algorithm}

\begin{algorithm}[h]
    \footnotesize
    \caption{FullRBnB: Rotation BnB without correspondences}
    \label{Alg3}
    \hspace*{\algorithmicindent} \textbf{Input:} \\
   \hspace*{\algorithmicindent}\hspace{1em} $\mathcal{X}_{lin} = \{(\bar{\textbf{d}}_{t_i},\tilde{\textbf{b}}_{t_i})\},\; (i=1,\cdots, N)$: Target embeddings\\
   \hspace*{\algorithmicindent}\hspace{1em} $\mathcal{Y}_{lin} = \{(\bar{\textbf{d}}_{s_j},\tilde{\textbf{b}}_{s_j})\},\; (j=1,\cdots, M)$: Source embeddings\\
    \hspace*{\algorithmicindent}\hspace{1em} $\mathcal{C}_\mathbf{R}$: Initial $SO(3)$ search cube\\
    \hspace*{\algorithmicindent}\hspace{1em} $\epsilon_\mathbf{R}$: BnB threshold 
    
\hspace*{\algorithmicindent}	\textbf{Output:}\\ \hspace*{\algorithmicindent}\hspace{1em} $\mathbf{R}^*$: Optimal rotation \\  
\hspace*{\algorithmicindent} \hspace{1em} $\mathcal{I}_f$: Resulted correspondences
    \begin{algorithmic}[1]
    \State $\mathcal{I}_f=\{\}$
    \State Add initial cube $\mathcal{C}_\mathbf{R}$ into priority queue $\mathbf{Q}_\mathbf{R}$
    \While{$\mathcal{I}_f.size < \bar{\nu}_r$}
    \State Read cube $\mathcal{C}_r$ with the greatest upper bound $\bar{\nu}_r$ from $\mathbf{Q}_\mathbf{R}$
    \For {\textbf{all} sub-cube $\mathcal{C}_{r_i}$}
    \State Compute the lower bound $\underline{\nu}_{r_i}$ \quad\text{\% \equref{equ:L2LLB}}
    \If {$\mathcal{I}_f.size < 2\underline{\nu}_{r_i}$}
    \State $\mathbf{R}^*$ $\leftarrow $ \texttt{LMOptimization}($\mathcal{X}_{lin}$, $\mathcal{Y}_{lin}$, $\mathbf{R}^*$)
    \EndIf
        \State Compute the upper bound $\bar{\nu}_{r_i}$ \quad\text{\% \equref{equ:L2LUB}}
        \If{$ \mathcal{I}_f.size <\bar{\nu}_{r_i}$} 
        \State Add $\mathcal{C}_{r_i}$ to queue $\mathbf{Q}_\mathbf{R}$
        \EndIf
    \EndFor
    \EndWhile
    \State $\mathcal{I}_f$ = \texttt{FindCorr}($\mathcal{X}_{lin}$, $\mathcal{Y}_{lin}$, $\mathbf{R}^*$)
        \State \textbf{return} $\mathcal{I}_{f}$, $\mathbf{R}^*$
    \end{algorithmic}
\end{algorithm}

\begin{algorithm}[h]
    \footnotesize
    \caption{TBnB: Translation BnB}
    \label{Alg4}
    \hspace*{\algorithmicindent} \textbf{Input:} \\
   \hspace*{\algorithmicindent}\hspace{1em} $\mathcal{X}_{lin} = \{(\bar{\textbf{d}}_{t_i},\tilde{\textbf{b}}_{t_i})\},\; (i=1,\cdots, N)$: Target embeddings\\
   \hspace*{\algorithmicindent}\hspace{1em} $\mathcal{Y}_{lin} = \{(\bar{\textbf{d}}_{s_j},\tilde{\textbf{b}}_{s_j})\},\; (j=1,\cdots, M)$: Source embeddings\\
    \hspace*{\algorithmicindent}\hspace{1em} $\mathcal{C}_\mathbf{t}$: Initial $\mathbb{R}^3$ search cube\\
    \hspace*{\algorithmicindent}\hspace{1em} $\epsilon_\mathbf{t}$: BnB threshold \\ 
    \hspace*{\algorithmicindent}\hspace{1em}
    $\mathcal{I}_f$: Correspondences
    
\hspace*{\algorithmicindent}	\textbf{Output:}\\ \hspace*{\algorithmicindent}\hspace{1em} $\mathbf{t}^*$: Optimal translation  
    \begin{algorithmic}[1]
    \State Set optimal error $e^*=+\infty$
    \State Add initial cube $\mathcal{C}_\mathbf{t}$ into priority queue $\mathbf{Q}_\mathbf{t}$
    \While{$e^*-\underline{e}_t<\epsilon_{\mathbf{t}}$}
    \State Read cube $\mathcal{C}_t$ with the lowest lower bound $\underline{e}_t$ from $\mathbf{Q}_\mathbf{t}$
    \For {\textbf{all} sub-cube $\mathcal{C}_{t_i}$}
    \State Compute the upper bound $\bar{e}_{t_i}$ \quad\text{\% \equref{equ:L2L_trans_UB}}
    \If {$\bar{e}_{t_i} < e^*$}
    \State ($e^*$, $\mathbf{t}^*$) $\leftarrow $ \texttt{LMOptimization}($\mathcal{X}_{lin}$, $\mathcal{Y}_{lin}$, $\mathcal{I}_f$, $\mathbf{R}^*$, $\mathbf{t}^*$, $e^*$ )
    \EndIf
        \State Compute the lower bound $\underline{e}_{t_i}$ \quad\text{\% \equref{equ:L2L_trans_LB}}
        \If{$\underline{e}_{t_i}< e^*$} 
        \State Add $\mathcal{C}_{t_i}$ to queue $\mathbf{Q}_\mathbf{t}$
        \EndIf
    \EndFor
    
    \EndWhile
            \State \textbf{return} $\mathbf{t}^*$
    \end{algorithmic}
\end{algorithm}
\newpage
\section{Analysis on Point-based and Parameter-based Methods}
\subsection{Measurement Variation of Point-based Registration}\label{appendix:PointVar}

In this section, we further analyze how the optimal rotation of point-based cost functions varies with changes in point location. For simplicity, our analysis focuses on 2-dimensional \textit{point-to-line} registration, considering only rotational transformations and assuming measurement points lie perfectly on the model shape without any noise.
In this case, the optimal rotation will initially be the identity element. We then examine how the optimal rotation changes as the noise in the points increases.

\begin{figure}[h]
    \centering
\includegraphics[width=0.5\columnwidth]{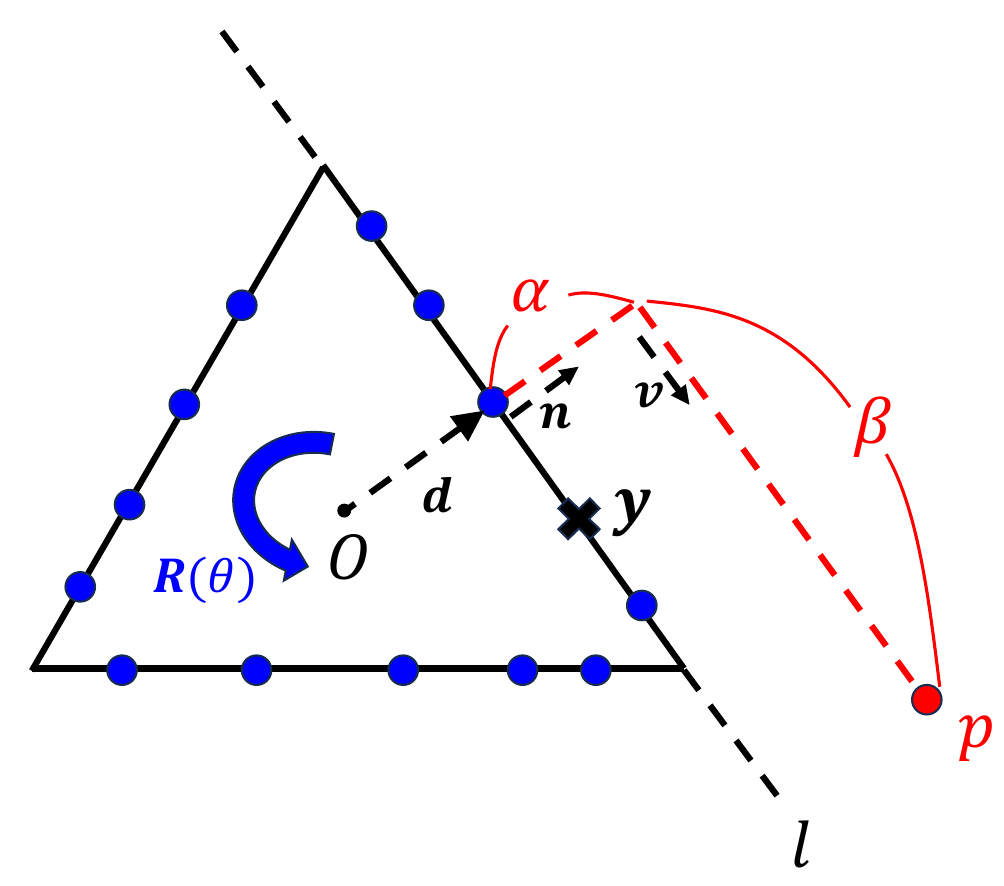}
\caption{\textbf{Measurement points on the model with outlier $\mathbf{p}$}. This figure illustrates measurement points without noise (\textcolor{blue}{blue}) lying on the corresponding model lines. When an outlier $\mathbf{p}$ (\textcolor{red}{red}), expressed by parameters $\alpha$ and $\beta$, is paired with line $l$, the optimal rotation $\mathbf{R}(\theta)$ may adjust to minimize the total \textit{point-to-line} costs.}\label{fig:point_analysis}
\end{figure} 
A measurement point on the model can be written as:
\begin{equation}
 \mathbf{p} = \mathbf{d} + \alpha\mathbf{n} + \beta\mathbf{v},  
\end{equation} where $\mathbf{d}$ is the displacement of the paired line with the point, $\mathbf{n}$ is the unit-norm displacement (normal vector), $\alpha$ is the noise in the normal direction, $\mathbf{v}$ is the unit-norm direction vector perpendicular to $\mathbf{n}$, and $\beta$ is the amount of displacement along $\mathbf{v}$. A rotation in $\mathbb{R}^2$ is expressed as an element of $SO(2)$ matrix:
\begin{equation}
    \mathbf{R}(\theta) = \begin{pmatrix}
\cos\theta & -\sin\theta \\
\sin\theta &  \cos\theta\\
\end{pmatrix}.
\end{equation} Therefore, as illustrated in \figref{fig:point_analysis}, a \textit{point-to-line} distance formulated by $\mathbf{p}$ is:
\begin{equation} \label{equ:d_p}
\mathbf{d}_\mathbf{p}=\left|\mathbf{n}^\top\left(\mathbf{R}(\theta)(\mathbf{d}+\alpha\mathbf{n} + \beta\mathbf{v}) - \mathbf{y}\right)\right|,
\end{equation} where $\mathbf{y}$ is the given point on the model. Denoting $\mathbf{n} = [n_x,n_y]^\top,\ \mathbf{d} = [d_x,d_y]^\top, \left\|\mathbf{d}\right\|=d,\ \ \mathbf{v} = [v_x,v_y]^\top$, this can be explicitly rewritten as:
\begin{align}
    \mathbf{d}_\mathbf{p} &= \left|\left(n_xd_x + n_yd_y + \alpha(n_x^2+n_y^2) + \beta(n_xv_x+n_yv_y)\right)\cos\theta + \left(\beta(n_yv_x - n_xv_y) + d_xn_y-n_xd_y\right)\sin\theta-\mathbf{n}^\top\mathbf{y}\right| \\ &= \left|(d+\alpha)\cos\theta +\beta\sin\theta-d\right| \label{equ:alpha_beta}.
\end{align}If $\alpha = 0$, indicating that $\mathbf{p}$ lies on the paired line, the minimizer $\theta$ of the cost is zero, regardless of the value of $\beta$. However, if $\alpha > 0$, we observe that absolute value of the minimizer $\theta$ decreases as $\beta$ increases. This aligns with our intuition: a point farther from the center can be slightly rotated to better fit the line. If such outlier points increase, the optimal rotation minimizing the sum of \equref{equ:d_p} deviates from the identity. In contrast, adding the number of points with smaller $\beta$ has no significant effect. This is because substantially increasing $\theta$ to align with near outliers would drastically increase the cost associated with inlier points, making the identity matrix the optimal rotation as before.

\subsection{Comparison of Curves on Manifold}\label{appendix:CurveCompare}
\hspace*{1em} From the paper, we observed that the sign ambiguity results in two distinct straight lines in the parameter space, which leads to suboptimal solutions. In this section, we investigate how these two lines are mapped as curves on the manifold and compare the lengths of these curves. First, we represent an element of the Grassmannian $\mathbb{A} \in \mathrm{Gr}(k,n)$ using its unique projection matrix, $\mathbf{P}_\mathbb{A} = \mathbf{AA}^\top$, where $\mathbf{A}$ is an orthonormal basis matrix \cite{bendokat2024grassmann}. 
Then, denoting two tangent vectors at $\mathbf{P} \in \mathrm{Gr}(k,n)$ as $\Delta_1, \Delta_2 \in T_\mathbf{P}\mathrm{Gr}(k,n)$, Riemannian metric at the tagent space is defined as:
\begin{equation}\label{equ:GrassMetric}
    g_\mathbf{P}(\Delta_1,\Delta_2) = \frac{1}{2}\mathrm{tr}(\Delta_1\Delta_2).
\end{equation}

Given two subspaces $\mathbb{A}, \mathbb{B} \in \mathrm{Gr}(k,n)$ and Riemannian metric of \equref{equ:GrassMetric}, minimal geodesic equation $\gamma(t)$ connecting two points ($\gamma(0) = \mathbf{P}_\mathbb{A}, \gamma(1) = \mathbf{P}_\mathbb{B}$) is derived as \cite{batzies2015geometric}: 
\begin{equation}\label{app:geodesicEqu}
    \gamma(t) = e^{t\mathbf{C}}\mathbf{P}_\mathbb{A}e^{-t\mathbf{C}},\; (\text{where }e^{2\mathbf{C}}=(\mathbf{I}-2\mathbf{P}_\mathbb{B})(\mathbf{I-2\mathbf{P}_\mathbb{A}})),
\end{equation}
Applying this result for connecting two embeddings of affine subspace $\mathbb{A}+\mathbf{c}, \mathbb{B} + \mathbf{d} \in \mathrm{Graff}(k,n)$, \equref{app:geodesicEqu} leads to $\gamma: [0,1] \rightarrow \mathrm{Gr}(k+1,n+1)$:
\begin{align}\label{app:geodesicAffine}
    &\gamma(t) = e^{t\mathbf{C}}\mathbf{P}_{z(\mathbb{A}+\mathbf{c})}e^{-t\mathbf{C}},\; (\text{where }e^{2\mathbf{C}}=(\mathbf{I}-2\mathbf{P}_{z(\mathbb{B}+\mathbf{d})})(\mathbf{I-2\mathbf{P}_{z(\mathbb{A}+\mathbf{c})}})),
\end{align}where the length of this geodesic is explicitly given by the root-sum-square of the principal angles between $\mathbf{Y}_{z(\mathbb{A}+\mathbf{c})}$ and $\mathbf{Y}_{z(\mathbb{B}+\mathbf{d})}$. 

We now derive the length of the curve mapped from the straight line in the Euclidean parameter space to the manifold. First, we define the mapping between the two spaces as $\phi(\cdot): \mathbb{R}^m \rightarrow \mathrm{Gr}(k+1, n+1)$, where $m$ is the dimension of the intermediate space (e.g., $m=4$ for a 3D plane parameter $(a, b, c, d)$). Then, the mapped curve from the Euclidean embedding is derived as $\phi(t):=\phi(\mathbf{v}(t))$, where $\mathbf{v}(t)$ is the straight line on the parameter space connecting two features represented as $\mathbf{v}_1$ and $\mathbf{v}_2$:
\begin{align}
    \mathbf{v}(t) = t\mathbf{v}_2 + (1-t)\mathbf{v}_1.
\end{align}
Then, the velocity at $t=t_k$ is derived by the chain rule:
\begin{align} \label{equ:chainrule}
    \dot{\phi}(t_k) &= (\frac{\partial\phi}{\partial\mathbf{v}}|_{\mathbf{v}=\mathbf{v}(t_k)})\cdot\dot{\mathbf{v}}(t_k) \\ &= (\frac{\partial\phi}{\partial\mathbf{v}}|_{\mathbf{v}=\mathbf{v}(t_k)})\cdot (\mathbf{v}_2 - \mathbf{v}_1).
\end{align} Since an element of Grassmannian is represented by the projection matrix $\phi(\mathbf{v}(t))\in \mathbb{R}^{(n+1)\times(n+1)}$, the partial derivative yields a 3D matrix $\frac{\partial\phi}{\partial\mathbf{v}} \in \mathrm{R}^{(n+1)\times(n+1)\times m}$. The notation $\cdot$ in \equref{equ:chainrule} denotes entry-wise multiplication, where ($i,j$) entry of $\dot{\phi}$ is:
\begin{equation}
\dot{\phi}_{ij} = \frac{\partial\phi_{ij}}{\partial\mathbf{v}} \cdot (\mathbf{v}_2 -\mathbf{v}_1)\end{equation}
Then, the length of the curve is approximated by uniformly discretizing $t\in[0,1]$ into $N$ samples:

\begin{align}\label{equ:ApproxLength}
    l\approx& \sum_{i=0}^{N-1} \sqrt{g_{\phi(i\Delta t)}(\dot{\phi}(i\Delta t),\dot{\phi}(i\Delta t))} \Delta t \\ =& \sum_{i=0}^{N-1} \frac{1}{\sqrt{2}}\mathrm{tr}(\dot{\phi}^2(i\Delta t))
 \Delta t, \; \text{ where } \Delta t = \frac{1}{N}.
 \end{align}

In the case of 2D lines, the embedding is represented by the coordinate $\mathbf{v} = (a,b,c)$ from the line equation $ax+by+c=0$. Then, the corresponding projection matrix $\phi(\mathbf{v})\in\mathbb{R}^{3\times3} $ in case of $(c<0)$ is:
\begin{align}\label{equ:ProjectionLine}
\phi(\mathbf{v}) = \frac{1}{a^2+b^2+c^2}\begin{bmatrix}
 b^2+c^2& -ab & ac \\
 -ab & c^2+a^2 & bc \\
 ac & bc & a^2+b^2 \\
\end{bmatrix}.
\end{align} Then, each entry of velocity is:

\begin{align}
    \dot{\phi}_{11} &= \frac{1}{(a^2+b^2+c^2)^2}\begin{bmatrix}
-2a(b^2+c^2) & 2a^2b & 2a^2c  \\
\end{bmatrix} \cdot \begin{bmatrix}
a_2-a_1 & b_2-b_1 & c_2-c_1 \\
\end{bmatrix}^\top \\ 
 \dot{\phi}_{12} &= \frac{1}{(a^2+b^2+c^2)^2}\begin{bmatrix}
-b(b^2+c^2-a^2) & -a(a^2+c^2-b^2) & 2abc  \\
\end{bmatrix} \cdot \begin{bmatrix}
a_2-a_1 & b_2-b_1 & c_2-c_1 \\
\end{bmatrix}^\top \\ 
\dot{\phi}_{13} &= \frac{1}{(a^2+b^2+c^2)^2}\begin{bmatrix}
c(b^2+c^2-a^2) & -2abc & a(a^2+b^2-c^2)  \\
\end{bmatrix} \cdot \begin{bmatrix}
a_2-a_1 & b_2-b_1 & c_2-c_1 \\
\end{bmatrix}^\top \\
\dot{\phi}_{21} &= \dot{\phi}_{12} \\
\dot{\phi}_{22} &= \frac{1}{(a^2+b^2+c^2)^2}\begin{bmatrix}
2ab^2 & -2b(a^2+c^2) & 2b^2c  \\
\end{bmatrix} \cdot \begin{bmatrix}
a_2-a_1 & b_2-b_1 & c_2-c_1 \\
\end{bmatrix}^\top \\ 
\dot{\phi}_{23} &=  \frac{1}{(a^2+b^2+c^2)^2}\begin{bmatrix}
-2abc & c(a^2+c^2-b^2) & b(a^2+b^2-c^2)  \\
\end{bmatrix} \cdot \begin{bmatrix}
a_2-a_1 & b_2-b_1 & c_2-c_1 \\
\end{bmatrix}^\top \\ 
\dot{\phi}_{31} &= \dot{\phi}_{13} \\ 
\dot{\phi}_{32} &= \dot{\phi}_{23} \\
\dot{\phi}_{33} &= \frac{1}{(a^2+b^2+c^2)^2}\begin{bmatrix}
2ac^2 & 2bc^2 & -2c(a^2+b^2)  \\
\end{bmatrix} \cdot \begin{bmatrix}
a_2-a_1 & b_2-b_1 & c_2-c_1 \\
\end{bmatrix}^\top
\end{align}

\begin{figure}[h]
    \centering
\includegraphics[width=0.7\columnwidth]{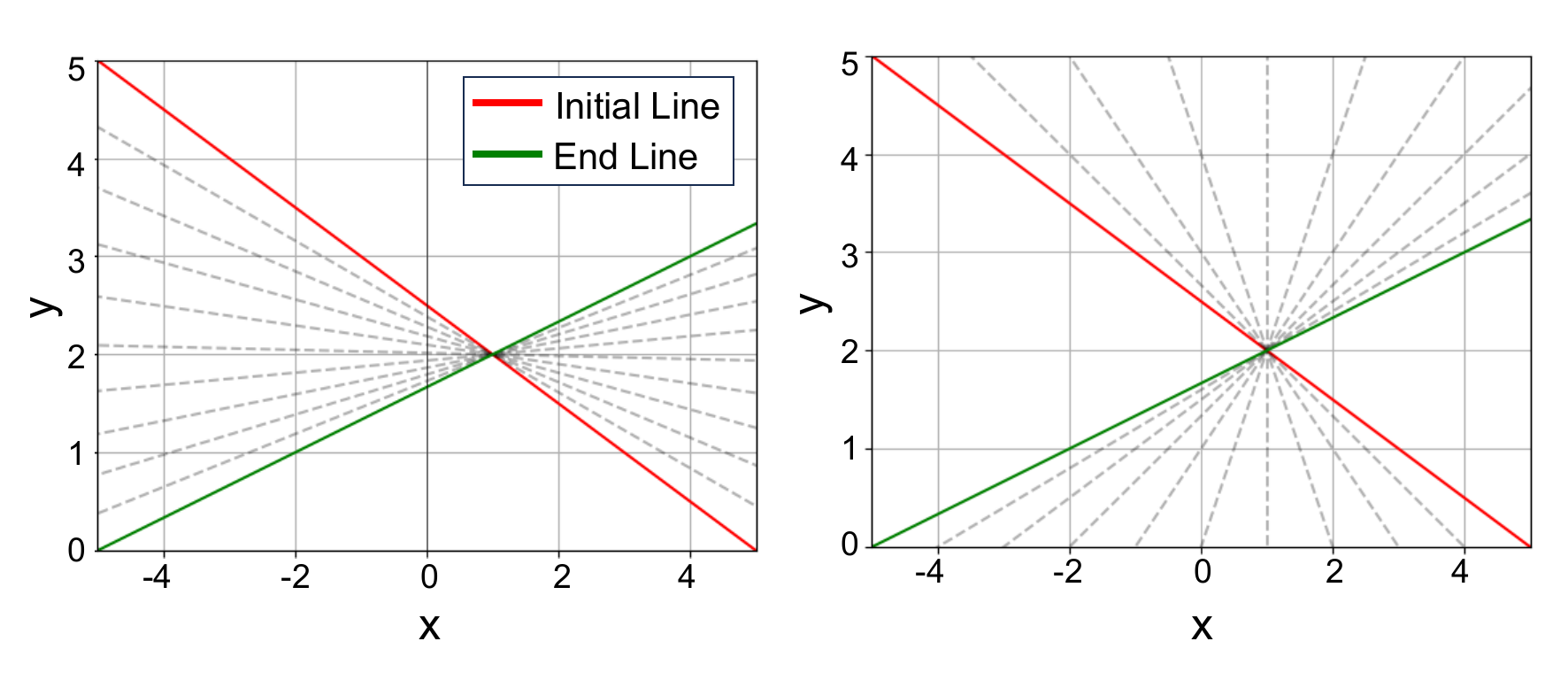}
    \caption{\textbf{Comparison of two projected straight lines in parameter space on the manifold.} This figure compares two curves connecting the initial line $\mathbf{v}_1 = (1, 2, -5)$ and the end line $\mathbf{v}_2 = (1, -3, 5)$. The left figure illustrates the projected trajectory of the straight line connecting ($\mathbf{v}_1$, $-\mathbf{v}_2$), while the right figure depicts the trajectory of the line connecting ($\mathbf{v}_1$, $\mathbf{v}_2$). }\label{fig:CurveCompare}
\end{figure} 

Given two lines $x+2y-5=0$ and $x-3y+5=0$ represented by $\mathbf{v}_1 = (1,2,-5)$ and $\mathbf{v}_2 = (1,-3,5)$, we numerically obtain the length of two curves which connect ($\mathbf{v}_1, \mathbf{v}_2$) and ($\mathbf{v}_1, -\mathbf{v}_2$). By defining two straight lines $\mathbf{c}_1(t) = t\mathbf{v}_2 + (1-t)\mathbf{v}_1$ and $\mathbf{c}_2(t) = -t\mathbf{v}_2 + (1-t)\mathbf{v}_1$ on the parameter space, the curve length projected on the manifold is obtained from \equref{equ:ApproxLength}. The trajectory of each curve is visualized as \figref{fig:CurveCompare}. From the results by selecting $N$ = 1000, the lengths of two curves, $l_1$ and $l_2$ are 2.7539 and 0.3876. This parameter-based approach inevitably selects a specific sign, resulting in different cost terms depending on the selection, as evidenced by their differing lengths. In the case of registering noisy data, this selective overweighing of a specific term may result in a suboptimal solution. In contrast, our cost function consistently minimizes the geodesic distance, effectively avoiding this ambiguity.

\begin{figure}[t]
    \centering
\includegraphics[width=0.7\columnwidth]{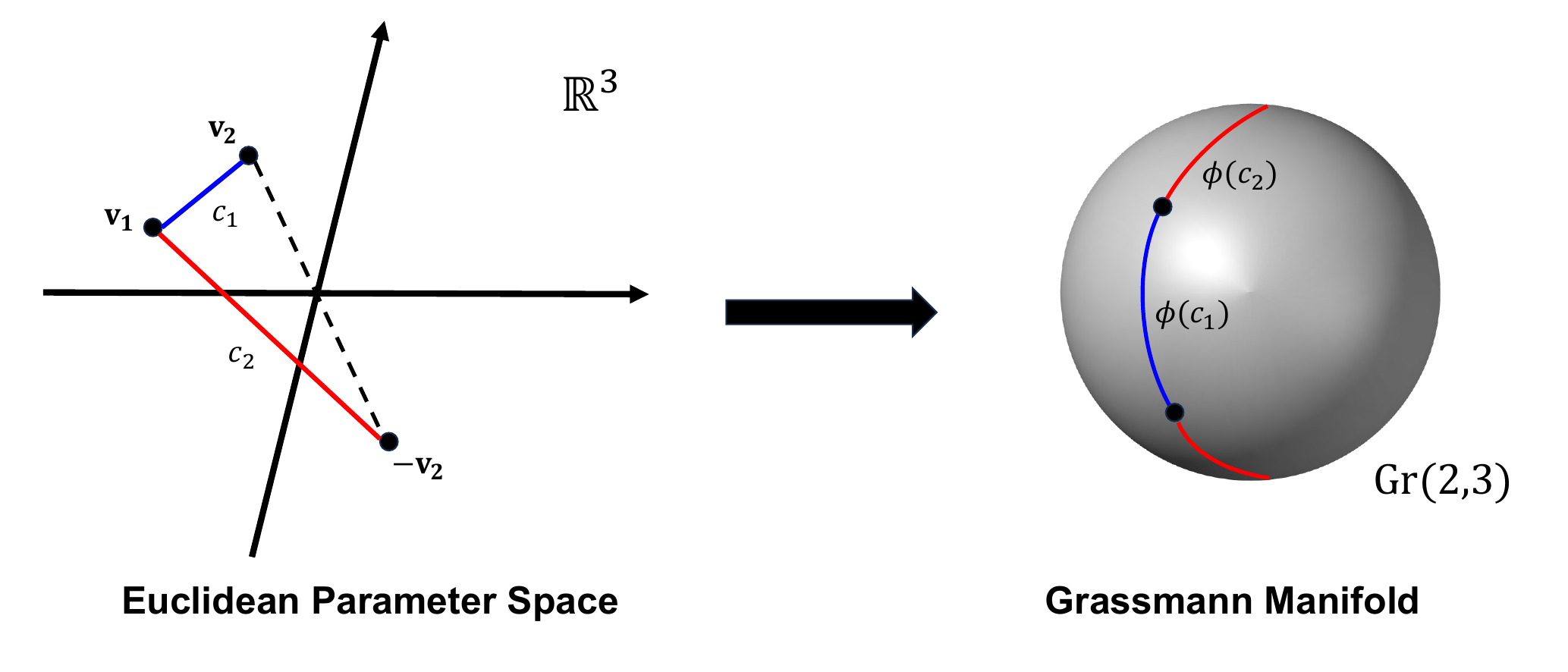}
    \caption{\textbf{Mapped straight lines on manifold.} Two distinct straight lines connecting the features, differing only by the sign of one parameter, may form a closed geodesic on the Grassmann manifold. Additionally, the projection of the shorter line (\textbf{\textcolor{blue}{blue}}) may align with the geodesic equation, while the longer line (\textbf{\textcolor{red}{red}}) is mapped to the longer arc.}\label{fig:closedGeo}
\end{figure} 

An interesting result is that, in every test case, one of the two lines consistently yielded the same length as the geodesic distance, matching to a precision of at least five decimal places and aligning with the trajectory of the geodesic in \equref{app:geodesicAffine}. Additionally, the sum of $l_1$ and $l_2$ always equaled $\pi$, which is twice the maximum geodesic distance of $\mathrm{Gr}(2,3)$, $\pi/2$. From this observation, we can infer that the projections of the two straight lines on the manifold are smoothly connected and form a closed geodesic on $\mathrm{Gr}(2,3)$ as illustrated in \figref{fig:closedGeo}. Assuming $\phi(\mathbf{v}(t))$  and \equref{app:geodesicEqu} represent identical curves on the manifold, and factorizing $\phi(\mathbf{v}(t))$ into the same form may reveal interesting properties of the Grassmann manifold and its representation as a projection matrix. This could also lead to an explicit representation of a longer geodesic connecting two points and a closed geodesic equation on the Grassmann manifold.

\section{Experiments Details}
\subsection{Time Complexity Analysis}\label{appendix:time}

In this section, we provide a computational time of experiments in \cref{sec:Experiments}. All the reported times represent the average time required to process a single set. For example, in the object registration experiment, the time for a specific outlier ratio is calculated by dividing the total time taken to compute its 500 sets by 500. 
\begin{figure}[h]
    \centering
    \begin{subfigure}{0.32\columnwidth}
        \includegraphics[width=\textwidth]{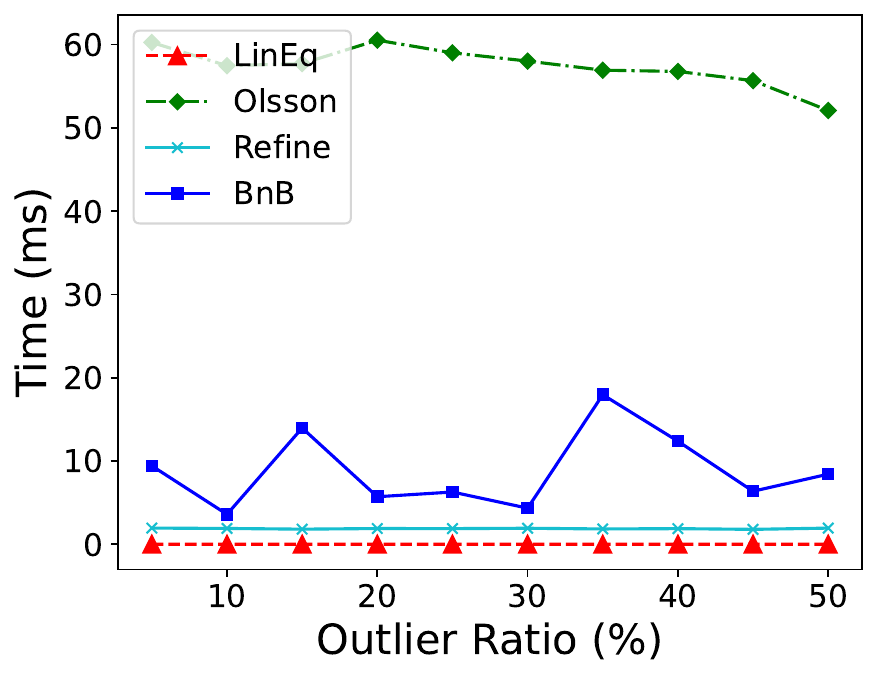}
        \caption{Object registration}
        \label{fig:objecttime}
    \end{subfigure}
    \begin{subfigure}{0.65\columnwidth}
        \includegraphics[width=\textwidth]{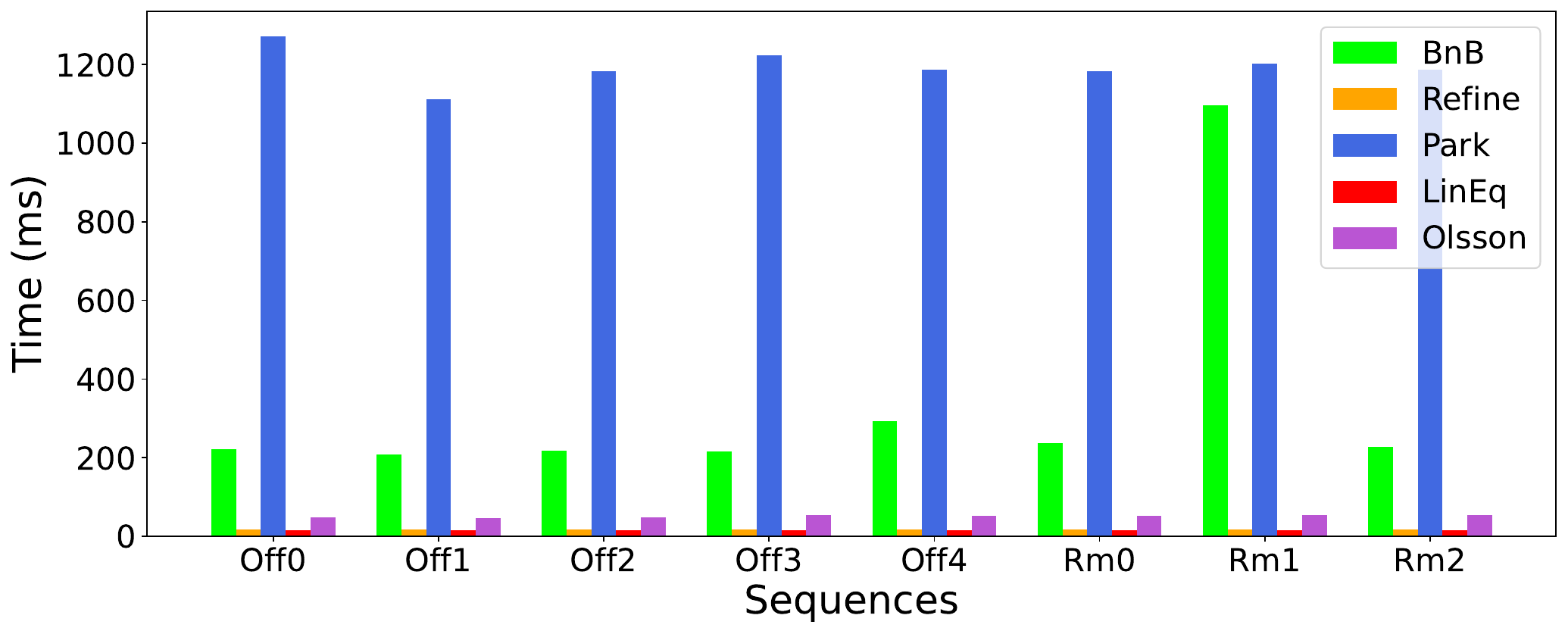}
        \caption{RGB-D odometry}
        \label{fig:odomtime}
\end{subfigure}
\caption{\textbf{Elapsed time analysis on object registration and RGB-D odometry tasks}. \subref{fig:objecttime} We examine the effect of the outlier ratio on the computational time for the object registration task. We show a comparable computational speed to the approximated parameter-based method while maintaining superior performance. \subref{fig:odomtime} Ours reported reasonable run-time speed for odometry on most sequences.}
\label{fig:time}
\end{figure} 

\textbf{Object registration task.} The result of the object registration experiment is illustrated in \figref{fig:objecttime}. Noticeably, compared to Olsson's method, which exploits every \textit{point-to-plane} correspondence, both our method and LinEq demonstrated shorter computation times by reformulating the original problem into a plane registration of 13 pairs. LinEq consistently achieved the shortest computation time because the algorithm derives its solution from two consecutive linear equations in the straightforward $\mathbf{Ax}=\mathbf{b}$ form. 


\textbf{RGB-D odometry task.} The elapsed time for each sequence in RGB-D odometry experiment is shown in \figref{fig:odomtime}. Overall, Park's method demonstrated the highest computational time, primarily due to the high resolution of the input images (1200 $\times$ 680) and the computationally intensive point cloud registration necessary to process a large number of points. Our BnB algorithm followed Park's method, showing its highest value in the \textit{Room1} sequence. As mentioned in \cref{Exp:RGBDodom}, this exceptionally high value was due to the challenging conditions of this sequence for line matching, as illustrated in \figref{fig:room1_fail}. However, compared to the failure of Pl\"{u}ckerNet and Olsson in this scenario, our BnB algorithm successfully estimated the optimal pose at the cost of significant computational time. 

\newpage

\begin{table}[h]
    \caption{\textbf{Elapsed time of 6D chessboard pose estimation using real images.} }\label{tab:PnL_time}
\centering
\begin{adjustbox}{width=0.7\linewidth}
\begin{tabular}{l|c|c|c|c|c|c|c}
\hline  
\multicolumn{1}{c|}{} & MinPnL \cite{RAL-2020-Zhou} & CvxPnL \cite{JMIV-2023-Agostinho}   & ASPnL \cite{PAMI-2016-Xu}    & ASP3L \cite{PAMI-2016-Xu}  & RoPnL \cite{RAL-2020-Liu}  & Refine & Ours   \\ \hline \hline t (ms)       &  80.91      &  36.34   & 0.51    & 0.62   &  278.18 & 1.27  & 1.24 \\ \hline
\end{tabular}
\end{adjustbox}
\end{table}

\begin{figure}[h]
    \centering
\includegraphics[width=0.7\columnwidth]{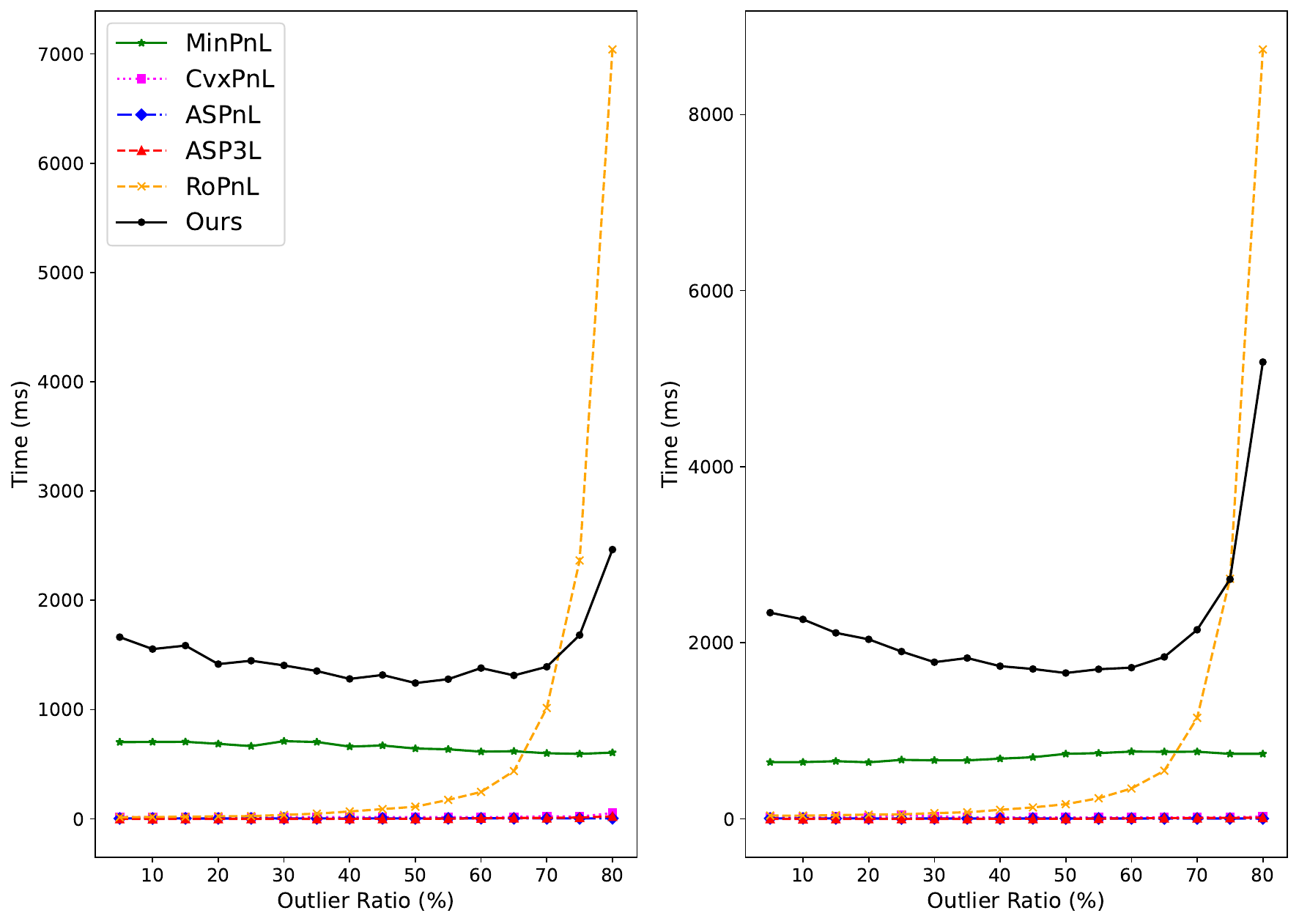}
    \caption{\textbf{Elapsed time of \ac{PnL} experiment with synthetic data}.}\label{fig:pnltime}
\end{figure}

\textbf{PnL task.} As shown in \figref{fig:pnltime}, in the synthetic data experiments, both our method and RoPnL showed an increase in computational time as the outlier ratio grew. Overall, RoPnL required less computational time than our method; however, its time increased exponentially with the outlier ratio, taking an average of over 7 seconds at an 80\% outlier ratio. 
In contrast, despite using BnB for calculating translation, our method achieved significantly shorter computation times at high outlier ratios and recorded much lower translational errors, as demonstrated in \cref{Exp:PnL}. 

In the chessboard experiment, RoPnL exhibited significantly large errors when using the same threshold as in the synthetic experiment, necessitating a reduction in the threshold for a fair comparison. This adjustment resulted in significantly higher computation times for RoPnL as shown in \tabref{tab:PnL_time}. Additionally, unlike the synthetic experiments that utilized 100 pairs, this real-world experiment employed only 7 pairs, corresponding to the 4$\times$3 chessboard pattern. As a result, we observed a significant reduction in our computation time, making it comparable to the results of ASPnL.

To summarize, as demonstrated in the results of Olsson in the object registration experiment and Park in the RGB-D odometry experiment, algorithms relying on points tend to require significantly more computational time as the number of measurements increases, in contrast to algorithms that compress this information into high-level features. Among these, parameter-based approaches such as LinEq and PlückerNet, which solve linear equations, achieved the shortest computation times but are limited by suboptimal solutions. Overall, our BnB algorithm successfully identified the inlier set and obtained an optimal solution in experiments involving a large number of pairs, particularly those with a high outlier ratio, but required significant computation time. In contrast, experiments with a smaller number of pairs demonstrated significantly shorter computation times. Considering this, the active use of plane features to robustly aggregate redundant lines, along with the selective use of meaningful features, is expected to alleviate these time complexity issues in practical applications.

\newpage
\subsection{Failure Cases in RGB-D Odometry}\label{appendix:OdomFail}
\begin{figure}[h]
    \centering
\includegraphics[width=0.8\columnwidth]{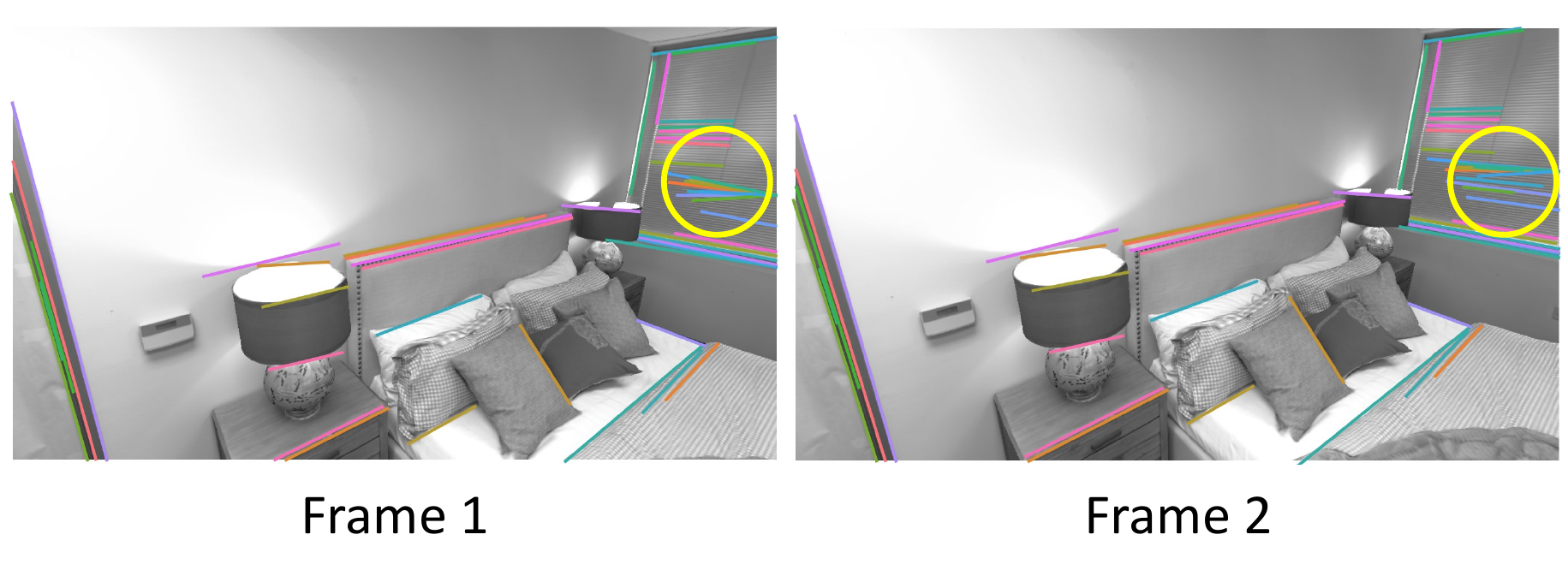}
    \caption{\textbf{Failure case of Olsson \cite{ICPR-2008-Olsson} in \textit{Room1} sequence.}}\label{fig:room1_fail}
\end{figure} 

This section presents visualizations of the frame pairs where each algorithm in the RGB-D odometry experiment recorded the largest error and includes a brief analysis of the causes. As shown in \figref{fig:room1_fail}, the areas highlighted with yellow circles reveal frequent failures in line segment matching using GlueStick \cite{ICCV-2023-Pautrat}. In the case of Olsson, which performs convex optimization by considering all correspondences, the algorithm was unable to handle such outliers internally, leading to a complete failure. Although PlückerNet mitigates this issue to some extent using RANSAC, it failed to find the optimal solution in this sequence where the outlier ratio is high. 

\begin{figure}[h]
    \centering
\includegraphics[width=0.8\columnwidth]{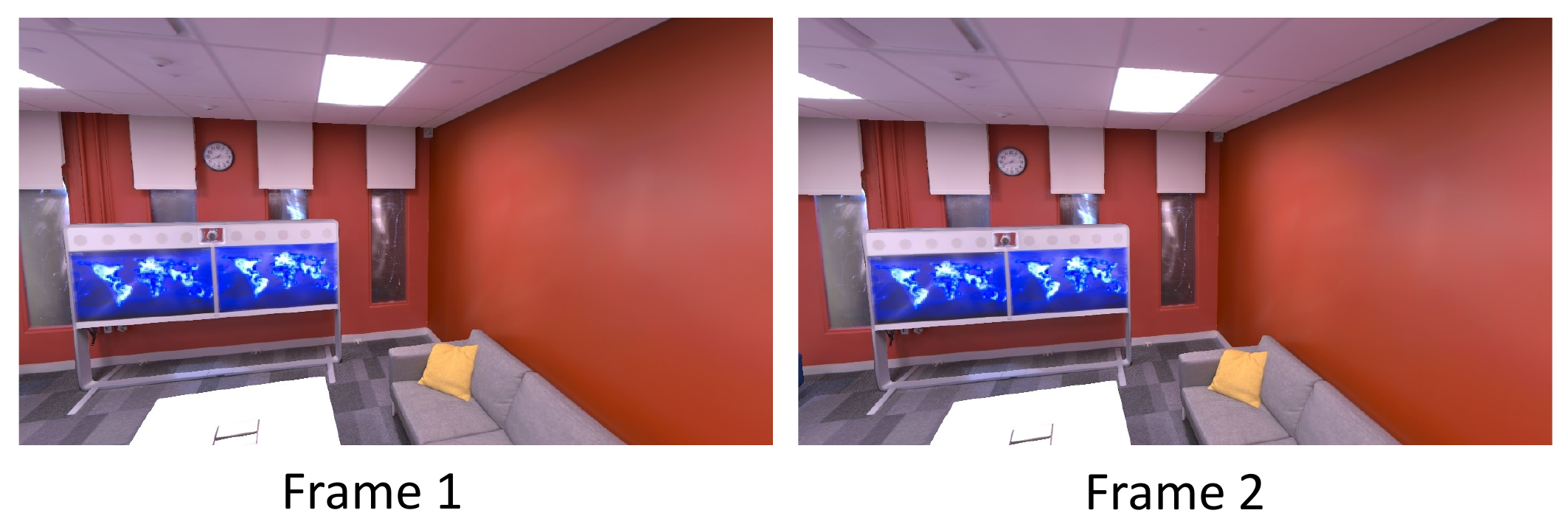}
    \caption{\textbf{Failure case of Park \cite{ICCV-2017-Park} in \textit{Office3} sequence }.}\label{fig:office3_fail}
\end{figure} 

Park generalizes the discrete intensity function obtained from the image into a continuous representation in 3D space by utilizing the gradient at a specific point. The gradient is estimated by minimizing the difference between the intensity of the continuous function and that of the discrete function across the neighborhood points. In scenes where the point cloud is dominated by points with the same intensity, as shown in \figref{fig:office3_fail}, the objective function exhibits minimal variation with changes in gradient values, leading to erroneous estimation. This gradient leads to incorrect calculations of the intensity function in these regions, resulting in large trajectory errors for the algorithm.

\end{document}